\documentclass[Afour,sageh,times]{sagej}

\usepackage{graphicx}%
\usepackage{multirow}%
\usepackage{rotating}
\usepackage{amsmath,amssymb,amsfonts}%
\usepackage{amsthm}%
\usepackage{mathrsfs}%
\usepackage[title]{appendix}%
\usepackage[dvipsnames]{xcolor}%
\usepackage{textcomp}%
\usepackage{manyfoot}%
\usepackage{booktabs}%
\usepackage{algorithm}%
\usepackage{algorithmicx}%
\usepackage{algpseudocode}%
\usepackage{listings}%
\usepackage{etoolbox}

\usepackage{bbding} % check ding
\usepackage{makecell,rotating} % table
\usepackage{caption}
\usepackage{pifont} % /makecell
\usepackage{dirtytalk} % /say
\usepackage[colorlinks,bookmarksopen,bookmarksnumbered,citecolor=black,urlcolor=black,allcolors={black}]{hyperref}
\usepackage{url} % href
\usepackage{arydshln} % dash hline

\usepackage{mfirstuc} % first upper case

\makeatletter
\renewcommand{\ref}[1]{%
	\@ifundefined{r@#1}%
	{X}%
	{\hyperref[#1]{\ref*{#1}}}%
}
\makeatother

\def \no{\textcolor{red}{\ding{55}}}
\def \yes{\ding{51}}
\def \na{\makecell{\centering \textcolor{gray}{\footnotesize n/a}}}

\def \frame{GOLLUM}
\def \framelong{G\MakeLowercase{rowable Online Locomotion Learning Under Multicondition}}
\def \framefull{\framelong{} (\frame)}

\def \overfor{overcoming catastrophic forgetting}
\def \expsim{exploitation of similarity}

\def \Duallearning{Dual Layer Learning Mechanism}
\def \duallearning{\MakeLowercase{\Duallearning}}

\def \layerstack{module}
\def \sup{supplementary document}

\def \columninterp{column-wise interpretation}
\def \layerinterp{layer-wise interpretation}
\def \init{direct knowledge transfer}
\def \behav{behavior model}
\def \pretrain{\rd{pretrain}}

\def \ZPG{C}
\def \ZPGfull{Sequential Central Pattern Generator}

\def \exp{experiment}

\def \videonormal{\href{https://youtu.be/MWWjpvYuwh0}{\url{https://youtu.be/MWWjpvYuwh0}}}
\def \videoslope{\href{https://youtu.be/znNi1mlLjEQ}{\url{https://youtu.be/znNi1mlLjEQ}}}
\def \videobroken{\href{https://youtu.be/HugIBO6cnNo}{\url{https://youtu.be/HugIBO6cnNo}}}
\def \videoterrain{\href{https://youtu.be/fGCy8CXPuO0}{\url{https://youtu.be/fGCy8CXPuO0}}}

\def \videodecompose{\href{https://youtu.be/PxAl___xCT8}{\url{https://youtu.be/PxAl___xCT8}}}
\def \videobehavmodel{\href{https://youtu.be/EGElrNx_kCE}{\url{https://youtu.be/EGElrNx_kCE}}}
\def \videofreq{\videonormal{} after 0:50 mins}
\def \videosim{\href{https://youtu.be/M6As_PgCDME}{\url{https://youtu.be/M6As_PgCDME}}}
\def \videoquad{\href{https://youtu.be/cmjijGxLLvA}{\url{https://youtu.be/cmjijGxLLvA}}}
\def \videobone{\href{https://youtu.be/qEqFoGwawpo}{\url{https://youtu.be/qEqFoGwawpo}}}
\def \videodemo{\href{https://youtu.be/mgONmN1hBwo\&t=34}{\url{https://youtu.be/mgONmN1hBwo\&t=34}}}
\def \videodemomorf{\href{https://youtu.be/fnWl33OQpak}{\url{https://youtu.be/fnWl33OQpak}}}

\def \gitlink{\href{https://github.com/Arthicha/GOLLUM.git}{\url{https://github.com/Arthicha/GOLLUM.git}}}

\def \firstsequence{on different slopes}
\def \secondseqeunce{on different slopes with potential motor dysfunction}
\def \thirdseqeunce{on different terrains}

\def \ttable{flat rigid floor}
\def \bluemat{thin mat}
\def \sponge{thick sponge}
\def \rough{rough paver}
\def \roughslope{inclined paver}
\def \gravels{gravel field}
\def \deg{$^\circ$}

\def \citetraining{\citep{MELA,dogrobot_teacherstudent,kaise_scirobotics,dogrobot_massivelyparallel,LILAC_latent_method,HlifeRL_optionbased,KeepLearning,hexapodlearning_fc_sixmodule,fourleg_cpg_multihead,droq,mathias_cpgrbf,cpgrl,aicpg,plastic_matching,dreamwaq,piggyback,dogrobot_teacherstudent_original,robotskillgraph,viability,identifyimportantsensor}}

\def \citefor{\citep{lifelong_hamburg,bio_liftlonglearning,HlifeRL_optionbased,review_continual,regularizationsuck}}

\def \figoverview{Figure~\ref{fig:overview}}
\def \figoverviewnet{\figoverview\sfig{a}}
\def \figoverviewhor{\figoverview\sfig{b}}
\def \figoverviewver{\figoverview\sfig{c}}

\def \figexpnormal{Figure~\ref{fig:expnormal}}

\def \fignormalsnap{Figure~\ref{fig:expnormal}\sfig{b}}

\def \figmethod{Figure~\ref{fig:method}}
\def \figcind{Figure~\ref{fig:cpgtwotypes}\sfig{a}}
\def \figcdep{Figure~\ref{fig:cpgtwotypes}\sfig{b}}
\def \figmetsignal{Figure~\ref{fig:method}}

\def \figprimaryvalue{Figure~\ref{fig:expprimarylearning}\sfig{a}}
\def \figprimaryobservation{Figure~\ref{fig:expprimarylearning}\sfig{b}}
\def \figprimarymatrix{Figure~\ref{fig:expprimarylearning}\sfig{c}}
\def \figprimaryskill{Figure~\ref{fig:expprimarylearning}\sfig{d}}

\newcounter{snum}
\newcounter{mnum}
\newcounter{rnum}

\newcommand{\review}[1]{{{#1}}}

\newcommand{\reviewfinal}[1]{{{{#1}}}}

\newcommand{\bd}[1]{\textbf{#1}}

\newcommand{\mrow}[2]{\multirow{#1}{*}{#2}}
\newcommand{\mcol}[2]{\multicolumn{#1}{c}{#2}}

\newcommand{\mc}[1]{\makecell{#1}}

\newcommand {\gry}[1]{\textcolor{gray}{\footnotesize #1}}
\newcommand {\rd}[1]{\textcolor{red}{#1}}

\newcommand {\sfig}[1]{\textbf{\small #1}}

\begin{document}

\runninghead{Interpretable Neural Control with Online Learning}

\title{Growable and Interpretable Neural Control with Online Continual Learning for Autonomous Lifelong Locomotion Learning Machines}

\author{Arthicha Srisuchinnawong\affilnum{1,2} and Poramate Manoonpong\affilnum{1,2}}

\affiliation{\affilnum{1}Embodied AI and Neurorobotics Laboratory, SDU Biorobotics, The M\ae rsk Mc-Kinney M\o ller Institute, The University of Southern Denmark, Odense, Denmark \\ \affilnum{2}Bio-Inspired Robotics and Neural Engineering Laboratory, School of Information Science and Technology, Vidyasirimedhi Institute of Science and Technology, Rayong, Thailand}

\corrauth{Poramate Manoonpong}

\email{poma@mmmi.sdu.dk, poramate.m@vistec.ac.th}

\begin{abstract}
	Continual locomotion learning faces four challenges: incomprehensibility, sample inefficiency, lack of knowledge exploitation, and catastrophic forgetting. Thus, this work introduces \MakeLowercase{\framelong{}} (\frame), which exploits the interpretability feature to address the aforementioned challenges. GOLLUM has two dimensions of interpretability: layer-wise interpretability for neural control function encoding and column-wise interpretability for robot skill encoding. With this interpretable control structure, GOLLUM utilizes neurogenesis to unsupervisely increment columns (ring-like networks); each column is trained separately to encode and maintain a specific primary robot skill. GOLLUM also transfers the parameters to new skills and supplements the learned combination of acquired skills through another neural mapping layer added (layer-wise) with online supplementary learning. On a physical hexapod robot, \frame{} successfully acquired multiple locomotion skills (e.g., walking, slope climbing, and bouncing) autonomously and continuously within an hour using a simple reward function. Furthermore, it demonstrated the capability of combining previous learned skills to facilitate the learning process of new skills while preventing catastrophic forgetting. Compared to state-of-the-art locomotion learning approaches, \frame{} is the only approach that addresses the four challenges above mentioned without human intervention. It also emphasizes the potential exploitation of interpretability to achieve autonomous lifelong learning machines.
\end{abstract}

\keywords{Continual Learning, Robot Learning, Explainable AI, Neural Control, Bio-inspired Robots, Central Pattern Generators}

\maketitle

\section{Introduction}
\refstepcounter{snum}

\label{sec:intro}

While animals can improve their locomotion skills throughout their lifetime \citep{bio_liftlonglearning}, robots are currently designed for certain predefined environment conditions and requires human involvement for pretraining in simulation \citetraining, providing task context \citep{HlifeRL_optionbased,KeepLearning}, designing leg encoding rules \citep{mathias_cpgrbf,mathias_nature}, and/or collecting sufficient fundamental skills/options \citep{robotskillgraph,hexapod_legmap,HlifeRL_optionbased}. This is owing to four key challenges \citep{bio_liftlonglearning,reviewInterpretableRL,lifelong_hamburg}: sample inefficiency, lack of knowledge exploitation, catastrophic forgetting, and incomprehensibility.

The first challenge exists even when robots experience a static environment. Given that most robots learn by trail-and-error using reinforcement learning \citep{RLbook} to maximize the reward feedback received from the interaction with the environment, learning requires a massive amount of training samples to estimate the reward gradient for stable policy update. This results in extensive training time, which ranges from 1 hour to 22 days \citetraining{}. For this reason, the majority of robot training occurs in accelerated simulations, which could later suffer from the reality gap due to simulation inaccuracy. This reality gap is a major problem that could affect robustness and performance \citep{dogrobot_massivelyparallel,dogrobot_teacherstudent}. Several methods, such as online system identification \citep{dogrobot_teacherstudent,dogrobot_teacherstudent_original}, have been proposed to mitigate the impact on performance; however, some performance gap still remains. An effective solution to achieve higher performance seems to be real-world fine tuning, which could add two extra hours \citep{KeepLearning}.

The second challenge emerges as soon as the second environment condition is introduced. To learn locomotion on different environment conditions, one approach is training different skills simultaneously \citep{dogrobot_massivelyparallel}. However, this takes a significant amount of time and yields lower performance \citep{SimultaneousVSIncremental,regularizationsuck}. Therefore, incremental training has been suggested as a more viable option \citep{HlifeRL_optionbased,SimultaneousVSIncremental,regularizationsuck}. Nevertheless, the worst-case scenario of training time when implementing incremental training is expected to grow proportionally to the number of conditions. Therefore, it is necessary to leverage the knowledge from one condition and efficiently apply it on the subsequent ones. One potential solution is to share the same parts of the network between different behaviors or skills \citep{MELA,mathias_nature}; however, a straightforward implementation could lead to the third challenge: catastrophic forgetting.

\def \replaynottoforget{\review{To tackle this problem, researchers have explored regularization-based techniques \citep{review_continuallearningtheory,forget_progresscompress,forget_regularization} and rehearsal-based techniques with experience replay \citep{review_continuallearningtheory,forget_largereplay,forget_selectivereplay,forget_mapreplay,forget_selfgenerate}. While the former techniques can be effective, they may still suffer from performance degradation due to improper regularization scaling. For example, \cite{forget_regularization} reported a 20\% decrease in performance using elastic weight consolidation. The latter techniques, on the other hand, can be computationally expensive. For example, the technique required 800k episodes of data when implemented straightforwardly \citep{forget_largereplay}. Additionally, this technique could suffer from performance degradation when the selected data \citep{forget_selectivereplay,forget_mapreplay} is insufficient or the generated data \citep{forget_selfgenerate} is inaccurate. \cite{forget_selfgenerate} reported 50\% performance degradation in such cases.} Therefore, the most efficient solution might be to completely separate different domains of knowledge from each other \citep{KeepLearning,BaysianIncremental,HlifeRL_optionbased}}

In the third challenge, catastrophic forgetting means that previous knowledge encoded in learned parameters (learned weights) of a neural control network may be replaced by the new knowledge (new weights) \citefor{} when experiencing new conditions. This can cause 5--70\% reduction in performance \citep{HlifeRL_optionbased,regularizationsuck} causing what has been learned to be forgotten. \replaynottoforget; yet, this approach brings us back to the second challenge of neglecting the \expsim. As a consequence, some roboticists tend to define a fixed working domain and then freeze the network after extensive training \citep{MELA,mathias_nature,dogrobot_massivelyparallel,BaysianIncremental}.

While the first three challenges remain unresolved, a fourth challenge emerges with the use of black-box models, such as large or deep neural networks. These models, along with their learning outcomes, are difficult to comprehend \citep{reviewInterpretableRL,review_mythos,review_stopexplain,taxonomiesXAI,darpaxai}. As a result, decomposing such a large black-box network into submodules or subnetworks for local functional analysis (decomposability), understanding its underlying learning process (transparency), simulating its working process (simulatability), verifying its results, and even making effective modifications all become highly challenging. Consequently, current legged robots are not only limited to a specific set of behaviors/environments defined at the design time but also lack the trustworthiness after extensive training.

\begin{figure*}%[!h] 
	\centering
	\includegraphics[width=0.98\textwidth]{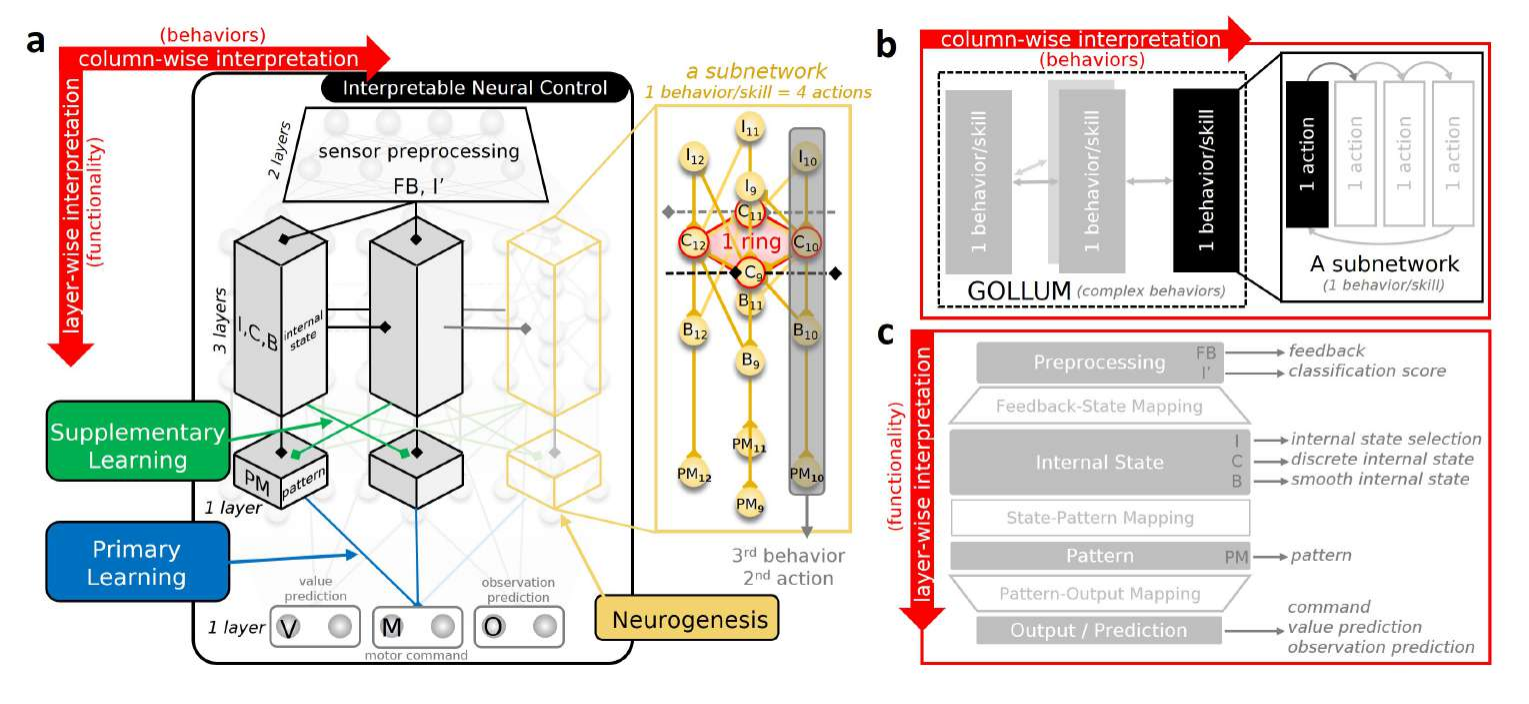}
	\caption{\sfig{a} \framefull{} consists of an interpretable neural control for motor command generation, a dual learning mechanism (primary learning for efficient locomotion learning and supplementary learning for exploiting shared skills), and a neurogenesis for implementing new skills. The interpretable neural control has two interpretation dimensions (column-wise and layer-wise). \sfig{b} In the horizontal/column-wise interpretation, neural columns are created by neurogenesis based on observation and value prediction mismatches. Each encodes a specific behavior/skill, which further includes multiple actions/target configurations. \sfig{c} In the vertical/layer-wise interpretation, four neural modules (sensory preprocessing, internal state, premotor/pattern, and motor/output modules) are stacked to fulfill network functionalities. The sensory preprocessing module is trained supervisedly on observation templates/predictions. The internal state module is precomputed and then fixed during the training. The premotor/pattern module is trained with the supplementary learning to exploit other learned behaviors/skills. The motor/output module is trained with the primary learning to refine/learn behaviors/skills. Combining horizontal and vertical interpretations, each interpretation coordinate thus represents a specific functionality at a specific action of a specific skill. For example, the neuron $\text{C}_\text{9}$ encodes the discrete internal state of the first action of the third skill. The neuron $\text{PM}_\text{10}$ encodes the pattern of the second action of the third skill, and the connection between $\text{PM}_\text{10}$ and an output encodes the output motor command (motor angle) for the second action in the third skill. A video with the neural visualization \citep{neurovis} is available at \videodecompose.}
	\label{fig:overview}
\end{figure*}

Previous studies tried to address all these challenges using progressively complex techniques. \review{These include i) architecture-based approaches that utilize complex rules to dynamically change the network structure and/or combine multiple trained networks \citep{review_continuallearningtheory,MELA,piggyback,forget_inteference,forget_progresscompress,robotskillgraph} and ii) representation-based approaches that employ complex processes to extract and maintain proper shared representations \citep{review_continuallearningtheory,HlifeRL_optionbased,dogrobot_teacherstudent_original,LILAC_latent_method}. In contrast, we propose here a less complex architecture-based approach designed based on interpretability.} In particular, this study hypothesizes that the interpretability introduced by two dimensions neural control and a \duallearning{} could not only provide understanding of the control system but also address other challenges, achieving lifelong learning locomotion intelligence through simpler mechanisms. Following this concept, we developed \MakeLowercase{\framelong{}} (\frame, \figoverview), which is a locomotion control and learning framework consisting of three key components: (1) an interpretable neural control network for motor command generation, observation prediction, and value prediction; (2) neurogenesis for incorporating new skills throughout operational lifetime; and (3) a \duallearning{} for fast and efficient learning without catastrophic forgetting, as summarized in \figoverview.

\section{Methods}
\refstepcounter{snum}
\label{sec:method}

\frame{} consists of three key components (\figoverview).
\begin{enumerate}
	\item An interpretable neural control is a neural network that maps sensory feedback/observation to motor commands and observation/value predictions. The motor commands are then used as the target positional trajectories for the robot, while the observation/value predictions are used during learning.
	\item A \duallearning{} is a learning mechanism that employs primary learning of motor command mapping connections to update the primary skills and employs the supplementary learning of inter-subnetwork connections to access other skills previously learned.
	\item A subnetwork neurogenesis is a mechanism that gradually creates new subnetworks when new conditions are detected through the deviation of the value and observation from predictions, providing new subnetworks/experts for encoding new skills.
\end{enumerate}

\subsection{Interpretable Neural Control} \refstepcounter{mnum} \label{sec:neuralcontrol}

\begin{figure*}
	\centering
	\includegraphics[width=0.98\textwidth]{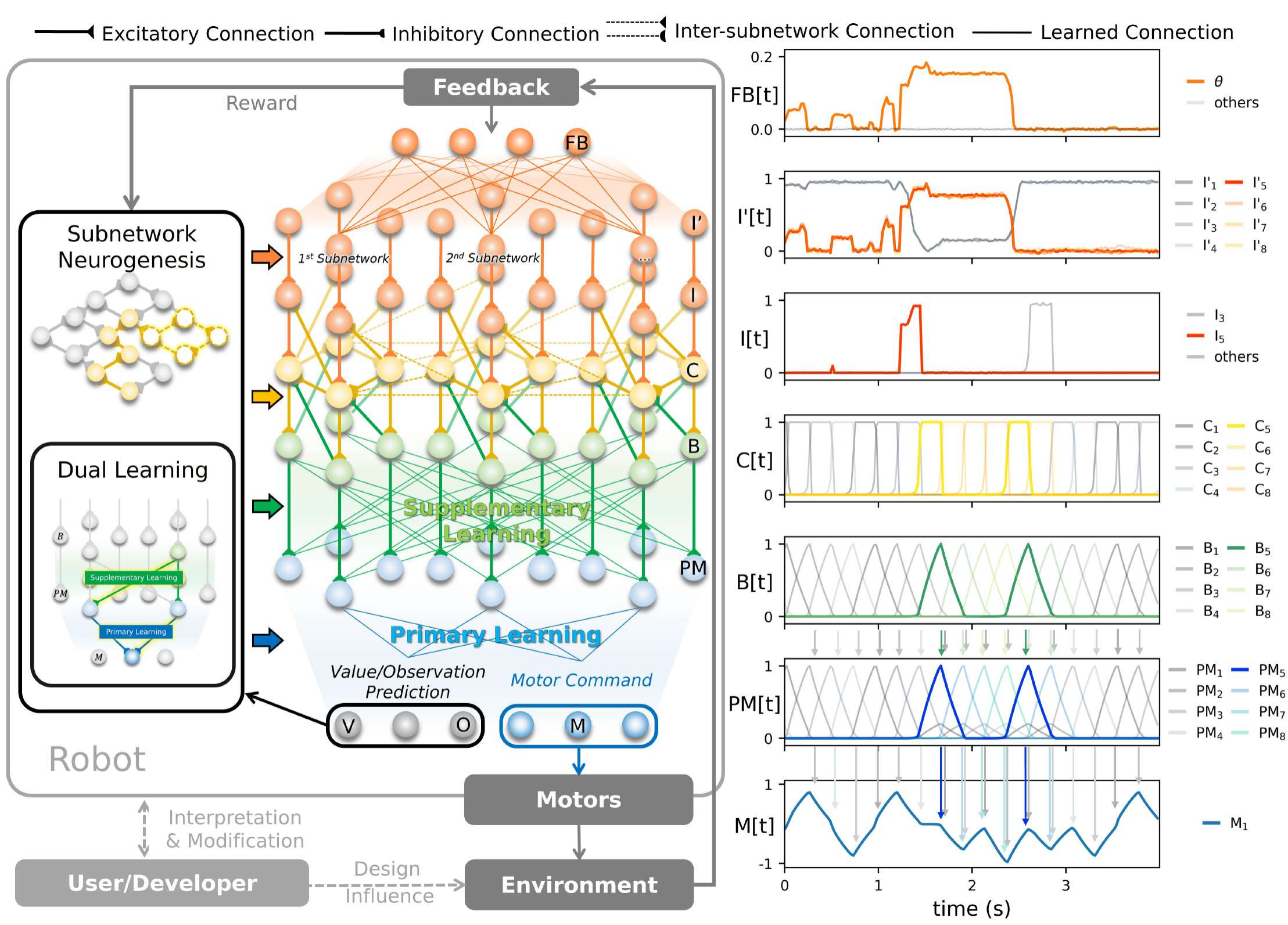}
	\caption{\frame{} framework, presented along with the corresponding neural activity signals: feedback ($FB[t]$, where $\theta$ denotes the robot pitch angle), first sensory preprocessing ($I'[t]$, classification score),  second sensory preprocessing ($I[t]$, internal state selection), sequential central pattern ($C[t]$, discrete internal state), basis ($B$, smooth internal state), premotor ($PM[t]$, pattern), and output ($M[t]$, motor command). The parameters of the network are summarized in the \sup. The signals from the first subnetwork are presented in gray scale, while those from the second subnetworks are in color. Receiving multiple feedback signals at the feedback layer (FB), the first sensory preprocessing layer (I') produces classification signals, which are later selected at the second sensory preprocessing layer (I) based on the activation of the sequential central pattern layer (C). The C layer produces two different sets of discrete internal states: the first group, in gray scale, activates at the start and after the selection signal $I_1[t]$; the second group, in yellowish, activates after the selection signal $I_4[t]$. The basis layer (B) then converts these discrete internal states to smooth internal states, forming the bases for shared action patterns at the premotor layer (PM). The action patterns are then projected to the outputs at the output layer (M, V, and O). The mapping from B to PM is trained by the supplementary learning to activate the proper patterns, while that from PM to M is trained by the primary learning to learn the proper action patterns. Upon encountering new conditions, indicated by a mismatch in value and observation predictions (V and O), the neurogenesis creates new subnetworks for learning new skills. Finally, due to the sparse neuron activation signals, the user/developer can gain insight into the network's processes and learned skills for further modify the results. A video demonstrating the mechanisms of \frame{} along with the neural visualization \citep{neurovis} is available at \videodecompose.}
	\label{fig:method}
\end{figure*}

The neural control is an interpretable neural network that maps sensory feedback/observation to motor command outputs: \review{a vector of $N_a$ independent action values, each of which is within the joint limits. $N_a$ is the number of motors ($N_a = 18$ for hexapod robot)}. The network is divided horizontally and vertically into different layers/subnetworks \citep{reviewInterpretableRL,review_stopexplain}, as shown in \figoverview. This results in two interpretation dimensions: \review{\columninterp{} (\figoverviewhor) and \layerinterp{} (\figoverviewver).}

\review{In the \columninterp{} (top view, \figoverviewnet{} and \figoverviewhor), the neural control consists of multiple columns/subnetworks, with ring neural networks \review{(i.e., groups of CPGs \citep{so2,cpgrl,mathias_cpgrbf,aicpg} that generate periodic outputs even in the absence of input signals)}, as highlighted in red in \figoverviewnet. Each column/subnetwork encodes a specific behavior/skill, constructed from multiple neural structures connected in a loop, representing various corresponding actions highlighted by the dark transparent box. In this work, four actions per subnetwork are used: two for the swing phase and two for the stance phase. The columns/subnetworks, connected via inter-subnetwork connections, allow transitions from one behavior/skill to another. Therefore, a neural activation within a subnetwork can be interpreted as the current condition/behavior and the corresponding actions (see \videodecompose). The inter-subnetwork connections reflect the self-organized structure of complex behaviors (i.e., \behav{} \citep{behaviorhierarchy1,behaviorhierarchy2}), as shown in \videobehavmodel.}

In the \layerinterp{} (side view, \figoverviewnet{} and \figoverviewver), the neural control comprises seven layers of \review{interpretable neural regressions \citep{taxonomiesXAI,reviewInterpretableRL,review_stopexplain}}. The layers are divided into four \layerstack s; each serves as a specific function as shown in \figoverviewver, providing interpretation transparency \citep{reviewInterpretableRL,review_stopexplain}. In this work, each layer is modeled as a discrete-time non-spiking \review{interpretable single neural regression layer \citep{reviewInterpretableRL,taxonomiesXAI}}, the activity of which is governed by
\begin{equation}
n_i[t+1] = f\left( \sum_j \left( w_{n_i,n_j} n_j[t]\right) + b_{n_i} \right),
\label{eq:neuron}
\end{equation}
where $n_i[t]$ denotes the activity of neuron $n_i$ at timestep $t$, $f()$ denotes the activation function, $w_{n_i,n_j}$ denotes the connection weights from $n_j$ to $n_i$, and $b_{n_i}$ denotes the bias of neuron $n_i$.

In the first \layerstack{} (\figoverview), the observations/sensory feedback signals are fed to sensory feedback neurons in \review{the sensory feedback layer (FB)}. \review{The first input preprocessing layer (I')} maps the sensory feedback to intermediate preprocessed inputs, acting as the subnetwork/skill classification scores. These classification scores are then used by the column structures/subnetworks distributed in the horizontal plane as the selection signals, choosing between those locomotion skills/behaviors. In the second \layerstack{} (\figoverview), \review{the second input preprocessing layer (I)} inhibits the classification signals that correspond to the inactive internal states, represented by the activities of \review{the \MakeLowercase{\ZPGfull{}} layer (\ZPG)}. Thus, only the classification signals that correspond to the active behaviors/active states activate, resulting in one hidden state and one action/configuration used at a time. In this layer, there are two types of \ZPG{} neurons: feedback-independent and feedback-dependent neurons. The former connects in loops forming multiple ring structures \review{(a ring structure is highlighted in red in \figoverviewnet)}, where the activation of one neuron always propagates to another, while the latter is employed for transition, where the activation in one ring structure transitions to the next when its input is presented \review{(this is indicated by the neuron with inter-subnetwork connections in \figoverviewnet)}. Thus, the connectivity in the layer determines the transition sequence/activation order of both the C neurons themselves and the ones in other layers. Subsequently, \review{the basis layer (B)} maps them to the corresponding \review{sparse} triangular shape bases, \review{which are later used to create the outputs}. The \review{sparse} bases are employed to ease learning with less correlated bases. In the third \layerstack{} (\figoverview), the \review{premotor layer (PM)} serves \review{as a layer of neurons encoding action patterns (i.e., sets of target joint configurations represented by the corresponding output mapping weights). This allows for the sharing of learned pattern/output mapping between different skills/subnetworks.} Finally, in the last \layerstack{} (\figoverview), \review{the output layer (M, V, and O)} maps the action patterns and bases to \review{motor positional commands (M)} for controlling the robot, \review{value predictions (V)} for the dual layer learning and new context identification by the neurogenesis, and \review{observation predictions (O)} serving as the observation template for learning the preprocessing \layerstack{} and context identification by neurogenesis. \review{The details are depicted in Figure~\ref{fig:method} as well as discussed layer by layer below}.

Given that the connectivity within the C layers determines the network structure, a connection matrix $\mathbf{\kappa}$ is used to parameterize the network. The element at row $r$, column $c$ ($\kappa_{rc}$) is a binary value representing the existence of the connection (i.e., transition) from neuron $C_r$ to $C_c$, which also determines the structure of other neurons as described in the following paragraphs. For each neuron $C_c$, decision making/path selection is not needed providing that it always activates after certain neurons ($\sum_r{\kappa_{rc}} = 1$). Thus, $C_c$ is a feedback-independent neuron, connecting in a loop within each ring-like structure/subnetwork. On the other hand, $C_c$ requires decision making/path selection if it activates after two or more. Thus, $C_k$ is a feedback-dependent neuron, linking ring-like structures/subnetworks.

\subsubsection{Sensory Feedback (FB)}

\begin{figure}[!h]
	\centering
	\includegraphics[width=0.98\linewidth]{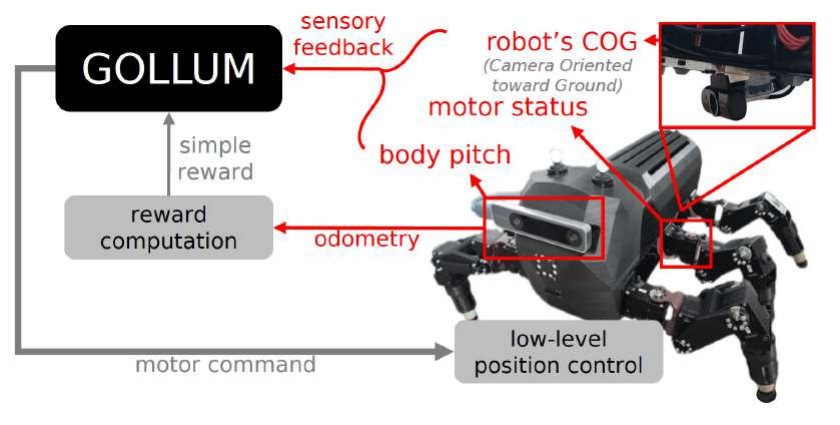}
	\caption{\review{MORF hexapod robot employed in this work presented long with its sensors, \frame{}, the robot interface, and the training process.}}
	\label{fig:robot}
\end{figure}

Four types of sensory feedback are provided to the robot as observations, namely, a body pitch feedback estimated from the tracking camera (Realsense T265) for slope detection, 18 signals of motor state feedback for motor dysfunction detection, the hue mean computed from the average hue pixels of images from the robot's COG for terrain color feature, and the hue standard deviation computed also from the standard deviation of hue pixels for terrain color range feature, as shown in Figure~\ref{fig:robot}. In total, the robot receives 21 dimensions of observation, that are passed to 21 sensory feedback neurons in the first layer. Note that other types of sensory feedback could be employed. 

\subsubsection{1$^{\text{st}}$ Input Preprocessing (I')}

The first input preprocessing layer (I') maps the sensory feedback/observation ($FB_k[t]$) to the intermediate preprocessed input ($I'_i[t]$), which represents the behavior/environment classification signals for activating the proper columns/subnetworks. The mapping is governed by \review{an interpretable regression model \citep{reviewInterpretableRL,taxonomiesXAI}}:
\begin{equation}
I'_i[t] = \sigma \left( \sum_k{\left( w_{I'_i,FB_k} FB_k[t] \right)+b_{I'_i}} \right),
\label{eq:ip1}
\end{equation}
\noindent where $\sigma()$ denotes sigmoid activation function. the parameters $w_{I'_i,FB_k}$ and $b_{I'_i}$ are \review{initially set to zero (i.e., no behavior selection) and then} trained supervisely to classify the corresponding subnetwork by minimizing the cross-entropy loss between the intermediate preprocessed inputs $I'_i[t]$ computed from $FB_k[t]$ and those computed from the feedback templates (i.e., observation prediction mapping weights, $FB_k[t] \leftarrow w_{O_i,B_k}$, described later).

Examples of the classification signals ($I'_i[t]$) are depicted in \figmetsignal, where two groups of classification signals can be observed. The classification signals of the first subnetwork ($I'_1[t]$--$I'_4[t]$) are higher when the robot body pitch signal ($\theta$) is below 0.2 in contrast to those of the second subnetwork ($I'_5[t]$--$I'_8[t]$) that are higher when the pitch signal is around 0.2 rad. Accordingly, these signals later activate the corresponding subnetworks.

\subsubsection{2$^{\text{nd}}$ Input Preprocessing (I)}

Receiving the classification signals, the second input preprocessing layer (I) blocks those that do not correspond to the current internal states and forwards the others to the next layer as the selection signals ($I_i[t]$). The layer is modeled as \review{an interpretable neural rule-based regression layer \citep{reviewInterpretableRL,taxonomiesXAI}}:

\begin{align}
I_i[t+1] =  \text{ReLU} \Bigl( \sum_k \left( w_{I_i,C_k} C_k[t]\right) + \tau_i I'_i[t] \notag \\ + (1-\tau_i) I_i[t] -1 \Bigr),
\label{eq:ip2}
\end{align}
where $\text{ReLU}()$ denotes the rectified linear unit activation function employed to scale the activities to be positive, and $w_{I_i,C_k}$ and $\tau_i$ denote the parameters. The first parameter, $w_{I_i,C_k}$, forms a sparse weight matrix, where it is set to 1 if $C_k$ is a feedback-dependent neuron and $\kappa_{ki} = 1$ (i.e., if there exists a connection from $C_i$ to $C_k$), and 0 otherwise. This setup selects only the classification signals required for behavior transition corresponding to the active internal states (Cs). The second and third parameters, $\tau_i$ determines the transition speed set to the same value as that in the basis layer, as also presented in the \sup. 

Examples of the selection signals ($I_i[t]$) are depicted in \figmetsignal, where merely two selection signals ($I_3[t]$ and $I_5[t]$) can be observed while the other classification signals observed in $I'[t]$ are blocked. The first signal ($I'_3[t]$) is forwarded to produce the classification signal $I_3[t]$ only when the discrete internal state $C_6[t]$ is active. Similarly, the second signal ($I'_5[t]$) is forwarded to produce the classification signal $I_5[t]$ only when the discrete internal state $C_4[t]$ is active. This later triggers the transition between the subnetworks.

\subsubsection{\ZPGfull{} (\ZPG)}

In this layer, series of two neurons with forward excitation and backward inhibition connections are connected in loops, forming ring structures. This structure is based on central pattern generator (CPG) in animal locomotion \citep{viability,fourleg_cpg_multihead}, neural locomotion circuit of \textit{Caenorhabditis elegans} \citep{nematodedecision,interpRL_dooperation}, and mushroom body of \textit{Drosophila melanogaster} \citep{bio_ringattractor,mushroombody}. The ring-like network structure serves as the main components for generating rhythmic patterns, which are then used as the internal states for forming repeated sequential actions, i.e., different locomotion patterns. Given the demonstrated rhythmic patterns in robot locomotion from previous studies \citetraining, \frame{} incorporates the rhythmic prior into its network structure, while maintaining a full action space. This enables the robot to learn diverse inter- and intra-leg coordination patterns.

Additionally, in this work, inter-connections between different rings are added, as shown in \figmethod. This allows activity patterns to propagate between columns/subnetworks, resulting in behavior transitions. To achieve such complex central pattern signals, two types of C neurons are designed: feedback-independent neurons (\figcind), which always allow activity propagation to the next neuron and generate rhythmic patterns, and feedback-dependent neurons (\figcdep), which allow activity propagation only when the corresponding selection signal from the second preprocessing layer (I) is provided, enabling behavior transitions.

\begin{figure}[!h]
	\centering
	\includegraphics[width=0.98\linewidth]{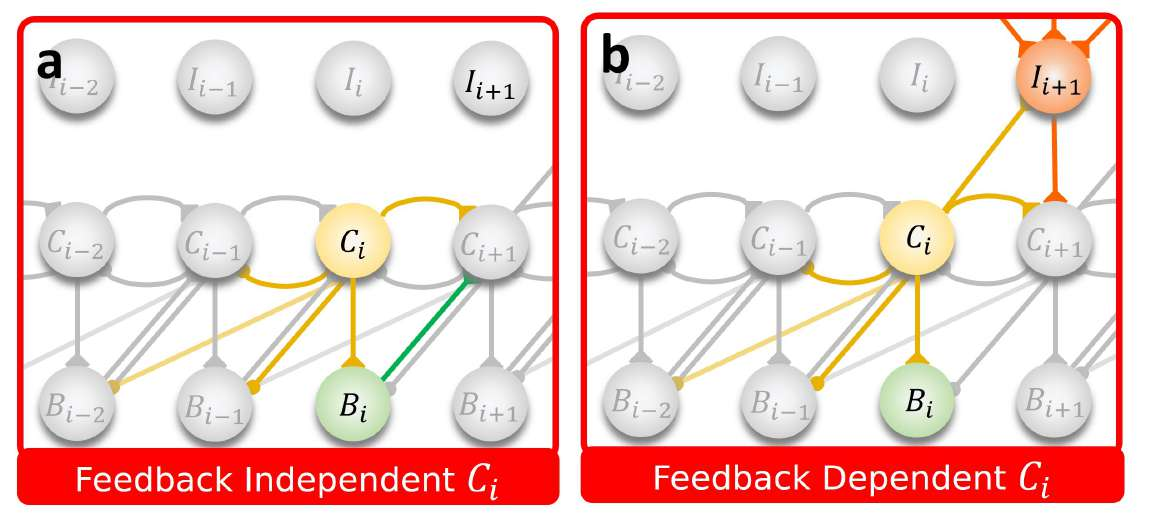}
	\caption{\sfig{a} Feedback independent and \sfig{b} feedback dependent sequential central pattern generator neurons ($C_i$). The former always propagates the activity of the former neuron ($C_{i-1}$) forward to the next ($C_{i+1}$) due to the excitatory connections from $C_i$ to $C_{i+1}$, $C_i$ to $B_i$, and from $B_i$ to $C_{i+1}$. The latter allows the propagation only when the selection input ($I_{i+1}$) is provided due to the excitatory connections from $C_i$ to $C_{i+1}$, from $C_i$ to $I_{i+1}$, and from $I_{i+1}$ to $C_{i+1}$.}
	\label{fig:cpgtwotypes}
\end{figure}

Despite having different functions, these two types are modeled as \review{an interpretable neural rule-based regression layer \citep{reviewInterpretableRL,taxonomiesXAI}} according to:
\begin{align}
C_i[t+1] = \sigma \Bigl( \sum_k \left( w_{C_i,C_k} C_k[t] + w_{C_i,B_k} B_k[t]\right) \Bigr. \notag \\+\Bigl. w_{C_i,I_i} I_i[t] +b_{C_i} \Bigr),
\label{eq:zpg}
\end{align}
where $\sigma()$ denotes sigmoid activation function. $w_{C_i,C_k}$, $w_{C_i,B_k}$, $w_{C_i,I_i}$, and $b_{C_i}$ are the parameters are set analytically as describe in Eqs.~\ref{eq:3zpg}--\ref{eq:condmatsolve} (also in Figures~\ref{fig:zpg_parameterselection} and \ref{fig:example_zpg_tb}) in the supplementary material so that the internal state activities propagate in the desired sequence, as shown in \figmetsignal{} and \videoslope. In this work, recurrent weight $w_{C_i,C_i}$ and bias $b_{C_i}$ are fixed to 20 and -13, respectively. For each $k$, the parameter $w_{C_i,C_k}$ is set to 7 if there exists a transition from $C_k$ to $C_i$ ($\kappa_{ki} = 1$) that enables forward propagation after $B_i$ or $I_{i+1}$ activates; however, it is set to -26 if there exists a transition from $C_i$ to $C_k$ ($\kappa_{ik} = 1$) that enables backward inhibition. To trigger the activity propagation of a feedback-dependent neuron, the parameter $w_{C_i,I_i}$ is set as 7, selecting the corresponding preprocess inputs as the trigger signal. By contrast, to trigger the activity propagation of a feedback-independent neuron, the parameter $w_{C_i,B_k}$ is set to 7 if there exists a transition from $C_k$ to $C_i$ ($\kappa_{ki} = 1$), thereby selecting the previous basis as the trigger signal. Thus, these parameters can be implemented as sparse matrices representing the structure of the behaviors, where most of the elements are zeros.

Example of the central pattern signals ($C_i[t]$) is depicted in \figmetsignal, where the activity of the feedback-dependent state $C_5[t]$ is triggered by the selection signal of the second behavior/skill ($I_5[t]$). After $C_5[t]$ is active, it triggers others feedback-independent states ($C_5[t]$--$C_8[t]$) until the selection signal of the first behavior/skill ($I_3[t]$) becomes active. Similarly, after $C_3[t]$ is active, it triggers other feedback-independent states in the first subnetworks ($C_1[t]$--$C_4[t]$). These discrete internal states are then smoothed to create the bases, which are later weight summed as the outputs.

\subsubsection{Basis (B)}
\label{sec:basis}

Taken the \ZPG{} activities as inputs, \review{the discrete internal states are smoothed and converted to triangular basis signals depicted in \figmethod. These triangular basis signals ($B_i[t]$) are passed through the premotor layer (PM) and finally linearly combined to produce the outputs. To produce triangular basis signals, this layer is modeled as an interpretable neural rule-based regression layer,} according to:

\begin{equation}
B_i[t+1] = \text{ReLU}\left( \sum_k{\left( w_{B_i,C_k} C_k[t] \right)}  + w_{B_i,B_i} B_i[t]  \right),
\label{eq:tb}
\end{equation}
\noindent where $\text{ReLU}()$ denotes the rectified linear unit function, employed to scale the bases to be between 0 and 1. The parameters in this layers are selected based on the neural dynamics of low-pass filter \citep{avis, zumogecko}. The parameters $w_{B_i,C_k}$ and $w_{B_i,B_i}$ are set to $\tau_i$, which denotes the propagation speed parameter of $C_i$. The parameters $w_{B_i,C_i}$ and $w_{B_i,B_i}$ are set to $\tau_i$ and $1-\tau_i$ for low-pass filtering of the sequential central patterns $C_i$. However, to refine and achieve triangular-shape bases, the parameters $w_{B_i,C_k}$ where $i \neq k$ are selected empirically as depicted in the \sup. If there exists a transition from $C_k$ to $C_i$ ($\kappa_{ki} = 1$), $w_{B_i,C_k}$ are set to $-0.5\tau_i$ and $w_{B_i,C_j}$ are set to $-0.25\tau_i$ for all the subsequent neurons $C_j$ ($\kappa_{ij} = 1$).

Examples of the basis signals ($B_i[t]$) are depicted in \figmetsignal, where the activities of sequential central pattern signals/discrete internal states ($C_i[t]$) are smoothed. This results in triangular signals, where each only intersects with its neighbor basis signals. For example, the basis $B_5[t]$ only activates when $B_4[t]$, $B_6[t]$, and $B_8[t]$ are non-zero. These triangular bases are used as the foundation, where their outputs are fed to the premotor layer (PM) to activate the corresponding action patterns and allow for the sharing of action patterns between different subnetworks (i.e., through different sets of bases). Subsequently, the signals are linearly combined to produce outputs.

\subsubsection{Premotor Layer (PM)}

To allow the sharing of the action patterns between different behaviors/skills, the premotor layer (PM) is added as an intermediate layer between the basis layer (B) and the output layer (M, V, and O). The activities of these neurons represent \review{shared action patterns (i.e., sets of target joint configurations)}. They can be accessed by other behaviors/skills \review{by simply activating the corresponding PM neurons at different desired levels} for \expsim{} between the learned behaviors/skills. This layer is modeled as an interpretable linear regression \citep{reviewInterpretableRL,taxonomiesXAI}, according to:
\begin{equation}
PM_i[t] = \sum_k{\left( w_{PM_i,B_k} B_k[t] \right)},
\label{eq:pmn}
\end{equation}
\noindent where $w_{PM_i,B_k}$ denotes the parameter of this layer. $w_{PM_i,B_i}$ is fixed to 1 to force the usage of the corresponding primary skill, while $w_{PM_i,B_k}$, where $PM_i$ and $B_k$ are in different subnetworks, are \review{initially set to zero (i.e., no skill sharing) and} then learned using supplementary learning (described later).

Examples of the premotor signals ($PM_i[t]$) are depicted in \figmetsignal, where the bases ($B_i[t]$) activate the action patterns with the same index ($PM_i[t]$), e.g., $B_5[t]$ activates $PM_5[t]$. Additionally, the second subnetwork is trained with the supplementary learning to exploit the action patterns from the first, resulting in the small supplementary action patterns of the first set ($PM_1[t]$--$PM_4[t]$, grayscale) being activated along with the action patterns of the second set ($PM_5[t]$--$PM_8[t]$, blues). These action patterns are then mapped to the outputs through the mapping connection weights trained by the primary learning.

\subsubsection{Output Layer (M, V, and O)}
\label{sec:met_output}

To produce the outputs of the nework, the output layer (M, V, and O) directly multiplies the activities of action patterns ($PM_k[t]$) or those of the bases ($B_k[t]$) before combining them. Given the sparse nature of the action patterns and bases, the mapping connection weights thus determine the corresponding output values. The connection weights $w_{M_j,PM_k}$ in Eq.~\ref{eq:mn} map the action patterns to the motor commands controlling the robot. The connection weights $w_{V,B_k}$ and $w_{V_\delta,B_k}$ in Eqs.~\ref{eq:vn}--\ref{eq:vcn} map the bases to the value prediction and its boundary, which are used for the learning and neurogenesis. The connection weights $w_{O_i,B_k}$ and $w_{O_{\delta i},B_k}$ in Eqs.~\ref{eq:on}--\ref{eq:ocn} map the bases to the observation/feedback predictions and their boundaries, which are used for the neurogenesis. This layer is thus modeled as three interpretable neural regressions (or an interpretable neural regression with three types of outputs), according to: 

%In this layer, there are three output types. First, the motor commands ($M_j[t]$) are the weighted summation of the high-level patterns ($PM_k[t]$), computed from
\begin{equation}
M_j[t] = \sum_k{\left( w_{M_j,PM_k} PM_k[t] \right)},
\label{eq:mn}
\end{equation}
\noindent where the parameters $w_{M_j,PM_k}$ are initially set to zero (i.e., starting at the default standing configuration) and learned using the primary learning (described later). At each timestep, the output $M_j[t]$ is used as the target position to control the robot.

%Second, the predicted value ($V[t]$) and its maximum deviation ($V_\delta[t]$) are estimated from the weighted summation of the bases, computed from

\begin{equation}
V[t] = \sum_k{w_{V,B_k} B_k[t] },
\label{eq:vn}
\end{equation}
\begin{equation}
V_\delta[t] = \text{max}\left( \epsilon_v \;,\; \sum_k{w_{V_\delta,B_k} B_k[t] } \right),
\label{eq:vcn}
\end{equation}
\noindent where $V[t]$ denotes the predicted value, $V_\delta[t]$ denotes the maximum deviation, $\text{max}()$ denotes the maximum function, and $\epsilon_v$ denotes an arbitrary small number, empirically set to 0.02 in this work. The parameters $w_{V,B_k}$ are initially set to zero (i.e., predicted value = 0.0) and then trained supervisedly with a high learning rate to minimize the square error of the value prediction ($\mathcal{L} = \Sigma_{t}(\Sigma V[t]-R[t])^2$), while the parameters $w_{V_\delta,B_k}$ are initially set to one (i.e., maximum prediction boundary) and then trained with a slow learning rate to reproduce the maximum deviation from the predicted value ($\mathcal{L} = \Sigma_{t}(V_{\delta}[t]-\max_t|R[t]- V[t]|)^2$). The learning rate is empirically selected to obtain the maximum final sum of rewards. Note that using a lower learning rate for divergence values aiming to further accelerate the learning has been found decreasing the learning stability and performance, as illustrated in the \sup. The predicted value ($V[t]$) and its maximum deviation ($V_\delta[t]$) are then used to compute the advantage during the learning and identify new environment conditions during the neurogenesis.

%Third, the predicted observation ($O_i$) and their maximum deviations ($O_{\delta i[t]}$) are estimated from the weighted summation of the bases, computed from
\begin{equation}
O_i[t] = \sum_k{w_{O_i,B_k} B_k[t] },
\label{eq:on}
\end{equation}
\begin{equation}
O_{\delta i}[t] = \text{max}\left( \epsilon_o \;,\; \sum_k{w_{O_{\delta i},B_k} B_k[t] } \right),
\label{eq:ocn}
\end{equation}
\noindent where $O_i[t]$ denotes the predicted $i^{th}$ observation/feedback, $O_{\delta i}[t]$ denotes the maximum deviation of the $i^{th}$ observation/feedback, $\text{max}()$ denotes the maximum function, and $\epsilon_o$ denotes an arbitrary small number,empirically set to 0.02 in this work. The parameters $w_{O_i,B_k}$ are initially set to zero (i.e., prediction observations = 0.0) and then trained supervisedly with a high learning rate to minimize the square error of the observation prediction ($\mathcal{L} = \Sigma_{i,t}(O_i[t]-FB_i[t])^2$), while the parameters $w_{O_{\delta i},B_k}$ are initially set to one (i.e., maximum prediction boundaries) and then trained with a low learning rate to reproduce the maximum deviation from the predictions ($\mathcal{L} = \Sigma_{i,t}(O_{\delta i}[t]-\max_t|FB_i[t]-O_i[t]|)^2$). The predicted observations ($O_i[t]$) and their maximum deviations ($O_{\delta i}[t]$) are then used to identify new environment conditions during the neurogenesis.

An example output signal ($M_1[t]$) is depicted in \figmetsignal, where each action pattern ($PM_i[t]$) is mapped to the corresponding action, i.e., each key point of $M_1[t]$. For example, $PM_4[t]$ (the primary action pattern) produces the lower peak of around -0.8 at 1 s and 3 s. An output command value of -0.8 is determined by the mapping connection weight from $PM_4$ to $M_1$. Similarly, $PM_5[t]$ (the primary action pattern) is scaled by the mapping weight and is combined with that from $PM_1[t]$ (the supplementary action pattern) to produce the output value of around 0.0 at 2.0 s and 2.5 s. As a result of having two sets of internal states ($C_1[t]$--$C_4[t]$ and $C_4[t]$--$C_8[t]$), the robot uses the first locomotion pattern in the period between 0.0 s and 1.5 s before switching to the second in the period between 1.5 s and 3.0 s and returning to the first after 3.0 s.

\subsection{\Duallearning}
 \refstepcounter{mnum}
\label{sec:duallearning}

The \duallearning{}, is a reward-based reinforcement learning algorithm that includes two learning types to exploit similarity and overcome catastrophic forgetting. First, the primary learning updates the motor command mapping connections (i.e., primary connections, from the action patterns (PM, blue, \figoverview) to the motor command (M, gray neurons, \figoverview)) to learn the primary skills (action patterns encoded in PM) corresponding to the active behavior while \overfor. Second, the supplementary learning updates the connections between subnetworks (i.e., supplementary connections, from the basis (B, green) to the action patterns (PM, blue)) to supplement the active primary skill with the exploitation of action patterns of other inactive behaviors without changing the primary connections.

\begin{figure}[!h]
	\centering
	\includegraphics[width=\linewidth]{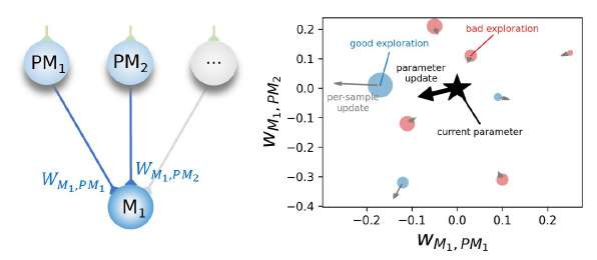}
	\caption{Visualization of the learning rule (Eq.~\ref{eq:rpg}) applied to the connection weights between two premotor neurons ($PM_{1}$ and $PM_{2}$) and a motor output ($M_1$), where the star denotes the coordinate of the current parameter values, blue dots denote the coordinates of the explored parameters with above-average returns (positive advantages), red dots denote the coordinates of the explored parameters with below-average returns (negative advantages), small grey arrows denote per-sample update gradients, and black arrow denotes the combined parameter update gradient applied to the connection weights. Note that, the size of the dots is proportional to the magnitude of the difference from the average, i.e., the advantages. Therefore, this visualization presents the working process of the learning rule that the update gradient is applied to move the parameters away from the bad explorations with fewer returns and toward the good explorations with higher returns.}
	\label{fig:learningplot}
\end{figure}

Both of the primary and supplementary connections are updated using the gradient-weighted policy gradient with consistent parameter-based exploration, modified from that reported in \cite{pgpe,pibb}, as described in Eqs.~\ref{eq:rpg} and \ref{eq:rpg_sigma}. The modified learning rule exploits the sparse basis signals by added weighting gain computed from the absolute of backpropagated gradient ($|\nabla_{\theta'_i} a'[t]|$) to down-weight/cancel out less relevant/non-relevant parameters and facilitate the learning. The primary learning updates are masked by the activation of the bases ($\mathcal{M}_i = 1 \; \text{if} \; (B_k > \epsilon), \; \text{else} \; 0$) to prevent the change of other primary skills, while the supplementary learning employs $\mathcal{M}_i = 1$.
\begin{equation}
\Delta \theta_i \approx 
\eta \; \mathcal{M}_i \sum_{\text{sample}} \sum_{t} |\nabla_{\theta'_i} a'[t]| \left( \frac{\theta'_i-\theta_i}{\sigma_i^2} \right) A[t],
\label{eq:rpg}
\end{equation}
\begin{equation}
\Delta \sigma_i \approx 
\eta \; \mathcal{M}_i \sum_{\text{sample}} \sum_{t} |\nabla_{\theta'_i} a'[t]| \left( \frac{(\theta'_i-\theta_i)^2 - \sigma_i^2}{\sigma_i^3} \right) A[t],
\label{eq:rpg_sigma}
\end{equation}
where $\Delta \theta_i$ denotes the update of parameter $\theta_i$, $\Delta \sigma_i$ denotes the update of exploration rate $\sigma_i$, $\eta$ denotes the learning rate, $t$ denotes time/timestep, $\nabla_{\theta'_i} a'[t]$ denotes the gradient of the explored action $a'[t]$ with respect to the explored parameter $\theta'_i$, $\sigma_i$ denotes the adaptable-exploration standard deviation of parameter $\theta_i$, and $A[t]$ denotes the standardized advantage estimate computed from the difference between the return and predicted value ($A[t] = \text{standardize}(R[t]- \Sigma V[t])$) at timestep $t$. The interpretation of this learning rule is presented in Figure~\ref{fig:learningplot}, where the update gradient (black arrow) is applied to move the parameters (star) away from the bad explorations with fewer returns (red dots) and toward the good explorations with higher returns (blue dots).

\subsubsection{Primary Skill Contribution}
\label{sec:primarycontribution}

Given that \frame{} uses one primary skill at a time for \overfor, the primary skill contribution is always (1, or 100\%) and can be represented by the subnetwork activities. The primary skill contribution or the activity of the subnetwork $\text{sub}_j$ ($\text{Pri. Skill}_{\text{sub}_j}$) is computed as follows
\begin{equation}
\text{Pri. Skill}_{\text{sub}_j} = \frac{\sum_{k \in \text{sub}_j}{B_k[t]}}{\sum_k{B_k[t]}},
\label{eq:pricontrib}
\end{equation}
where $\sum_{k \in \text{sub}_j}{B_k[t]}$ denotes the summation of all the bases corresponding to the subnetwork $\text{sub}_j$, and $\sum_k{B_k[t]}$ denotes the summation of all the bases.

\subsubsection{Supplementary Skill Contribution}
\label{sec:supplementarycontribution}

Given that the supplementary learning learns the combination ratio between the skills/action patterns from all subnetworks, the supplementary skill contribution of subnetwork $\text{sub}_j$ ($\text{Sup. Skill}_{\text{sub}_j}$) is computed as follows
\begin{equation}
\text{Sup. Skill}_{\text{sub}_j} = \frac{\sum_{j \in \text{sub}_j} \sum_k{ \left|w_{PM_{j},B_k}  B_k \right|}}{\sum_{j} \sum_k{ \left|w_{PM_{j},B_k}  B_k \right|}},
\label{eq:supcontrib}
\end{equation}
where $\sum_{j \in \text{sub}_j} \sum_k{ \left|w_{PM_{j},B_k}  B_k \right|}$ denotes the absolute summation of all $w_{PM_{j},B_k}  B_k$, where $j$ corresponds to the subnetwork $\text{sub}_j$, and $\sum_{j} \sum_k{ \left|w_{PM_{j},B_k}  B_k \right|}$ denotes the absolute summation of all $w_{PM_{j},B_k}  B_k$.

\begin{figure*}[!h] 
	\centering
	\includegraphics[width=0.95\textwidth]{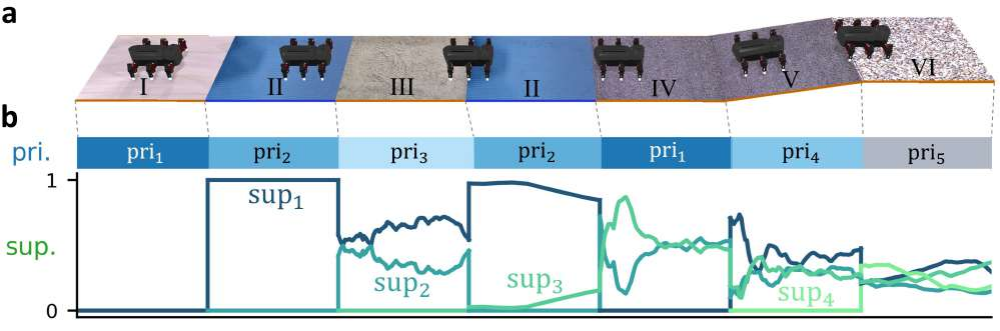}
	\caption{\sfig{a} Graphical illustration of locomotion learning \thirdseqeunce: (I) \ttable, (II) \bluemat{} (soft/deformable terrain), (III) \sponge{} (highly soft/highly deformable terrain), (IV) \rough, (V) \roughslope, and (VI) \gravels. \sfig{b} Corresponding primary skill contribution (i.e., the activation of the subnetworks/bases (B)) and supplementary skill contribution (i.e., the magnitude ratio of the inter-subnetwork connection from the active hidden states/bases (B) to the action patterns encoded in premotor neurons of other inactive subnetworks (PM)).}
	\label{fig:duallearning}
\end{figure*}

Figure~\ref{fig:duallearning} presents an example of online continual locomotion learning on different terrains, where the robot uses incremented primary skills (Eq.~\ref{eq:pricontrib}), presented by the activation of B, while exploiting the evolving ratio of supplementary skills (Eq.~\ref{eq:supcontrib}), presented by the evolution of the magnitude of the weights between the B and PM.

\subsection{Subnetwork Neurogenesis}
 \refstepcounter{mnum}
 \label{sec:neurogensis}
 
To deal with newly introduce conditions, the neurogenesis is activated to create new subnetworks/behaviors by modifying the Boolean connection matrix $\kappa$ that parameterizes the structure of the network. Taken the advantage of highly structure nature of the interpretable neural control, new columns/subnetworks, each of which has only few parameters ($\approx$ 200 sparse parameters that can be further compressed \citep{sparsecompress_blockwise,sparsecompress_viterbi}), can be created and added while slightly increasing memory and computational resources, as presented and discussed in the \sup. Inspired by the biological neuromodulators that are released as a result of uncertainty and surprise \citep{acl1}, new conditions are detected by the deviation in both the value (if $R[t] < \left( V[t] - V_\delta[t] \right) $) and observation (if there exists $| FB_i[t] - O[t] | >  O_{\delta i}[t]$, for $i$ in the number of sensory feedback signals). The neurogenesis is thus controlled by the embodied interaction with the environment. As the network grows and the robot learns more skills, it attempts to reuse a previously learned skill that has the most similar feedback pattern. The neurogenesis will occur only if the feedback received is different from the expected feedback while the skill receives less reward than expected.

After the detection and creation of the new subnetworks, they are employed for learning the new primary skills. Given that the robot autonomously switches to other behaviors featuring the most similar observation for realization of behavior transition, the weights from the previously active subnetwork are copied to the new subnetworks, thereby implementing the replication of the one with most similar feedback. This mechanism contributes to the \expsim{} during \init. 

\subsection{Experimental Robotics Platform}  \refstepcounter{mnum} \label{sec:robot}

In this work, a hexapod robot (Modular Robot Framework (MORF), \cite{morf}), shown in Figure~\ref{fig:robot}, is employed as the experimental platform. The robot has six legs, denoted as LF (left front leg), LM (left middle leg), LH (left hind leg), RF (right front leg), RM (right middle leg), RH (right hind leg). Each leg consists of three joints, denoted as number 1 (first joint, body-coxa joint), 2 (second joint, coxa-femur joint), and 3 (third joint, femur-tibia joint). There is a total of 18 active revolute joints, controlled by 18 XM430-W350-R Dynamixel motors with embedded positional sensors for low-level position control, torque sensors for reward calculation, and operating state feedback for broken motor/joint detection. The robot also includes an Intel RealSense tracking camera (T265) for odometry estimation, and a COG for obtaining terrain color features \citep{imageprocessing}. For simplicity, this work employs hue channel mean and standard deviation as inspired by HSI color space in image processing \citep{imageprocessing} and variational autoencoder \citep{vae}. Due to similar concept, other types of pre-trained latent variables could also be used, e.g., the latent space from pre-trained autoencoder \citep{vae}. In total, the robot weights approximately 4.7 kg.

To control the robot, motor target position commands are generated from a neural controller implemented on an external computer. In this experimental setup, the computer was an Intel Core i7 8750H CPU and Nvidia GeForce GTX 1050, and the motor position commands were generated at 20 Hz. The generated commands were sent via ROS wifi network to an onboard Intel NUC board (NUC717DNBE), serving as an onboard controller passing the target position commands to a standard low-level controller embedded in each motor via a U2D2 motor interface. Upon receiving the target positions, the Dynamixel low-level controller computes the velocity profiles with a maximum speed of 23 $rad/s$ and follows the profile with the default P-controller and $K_p = 800$ Dynamixel unit.

\section{Experiments and Results}
\refstepcounter{snum}
\label{sec:result}

To evaluate the performance of \frame, four locomotion learning \exp s were performed on MORF robot \citep{morf}. The first \exp{} investigated primitive locomotion learning in terms of sample efficiency on a regular flat terrain, while the next three \exp s shifted the focus toward continual energy-efficient locomotion learning in \firstsequence, \secondseqeunce, and \thirdseqeunce. 

In each \exp, the network was updated every single episode, using the trajectory from a short episodic experience replay of $N$ previous episodes ($N$ = 8, as in \cite{mathias_cpgrbf} and \cite{mathias_nature}). Each episode took 30 timesteps, being equivalent to approximately 1 gait cycle or 5 s.The hyperparameters of the training are summarized in the \sup. Owing to the limitation of the testing area, in all the \exp s, the robot was halted when the end of the testing area was reached and kept returning to the starting point until stable locomotion was obtained, i.e., until no further improvement was observed. The return $R[t]$ is computed from two types of single-term simple reward functions ($r[t]$) to demonstrate the locomotion learning with \frame{} under simple reward functions and remove the process of tuning multiple gains \citep{dogrobot_teacherstudent,kaise_scirobotics,mathias_cpgrbf}.

In the first \exp{} on primitive locomotion learning on regular terrain, the objective is to compare the learning efficiency and performance against the previous works in terms of speed-based reward function on regular flat terrain \citetraining. Therefore, the reward function $r[\tau]$ is defined as the forward speed $v[\tau]$ estimated from the robot odometry, as shown in Eq.~\ref{eq:rewardsimple}. After that, the return $R[t]$ is computed from the summation of the future reward over the horizon $H$, which is set as twice each basis activation time (14 timesteps) given that each basis signal overlaps (i.e., has influences) over its neighbors, as shown in Eq.~\ref{eq:speedrwd}.

\begin{equation}
r[\tau]= v[\tau],
\label{eq:rewardsimple}
\end{equation}
\begin{equation}
R[t]= \sum_{\tau = t}^{t+H} r[\tau].
\label{eq:speedrwd}
\end{equation}

In the following \exp s on continual locomotion learning, the objective is to study \frame{} under different environment conditions. The \exp s began under a variation of one environmental feature (i.e., different slopes) to \review{investigate the robot's ability to overcome catastrophic forgetting when encountering slopes beyond its hardware limit}. This was then extended to two environmental features (i.e., slopes and potential motor dysfunction) and their combination, exploring how the robot could exploit similarity between conditions (i.e., combining locomotion skills for a slope and motor dysfunction to tackle motor dysfunction on a slope). Finally, a more abstract and realistic example was demonstrated using real terrains. Given that the hexapod robot can achieve similar walking speed when traversing different terrain conditions \citep{pbird_hormone,pbird_morfendocrine,hexapod_optimize_cot}, using the speed reward in Eq.~\ref{eq:rewardsimple} can produce similar values across various terrain conditions. Thus, in the second, third, and fourth \exp s (i.e., continual learning under multiple conditions), the reward function was changed to the inverse cost of transport ($\text{COT}[\tau]$), depending on both the speed and energy consumption, as shown in Eq.~\ref{eq:cot}. This energy-related evaluation function has been shown to vary across multiple terrain conditions \citep{pbird_hormone,pbird_morfendocrine,hexapod_optimize_cot}, thus ensuring different optimal behaviors under different conditions and increasing the complexity of locomotion learning. The return is then computed from the summation of reward over the same horizon ($H$ = 14) to emphasize optimizing the minimum performance, as shown in Eq.~\ref{eq:cotrwd}.

\begin{equation}
r[t] = \frac{1}{\text{COT}[\tau]} = \frac{mgv[t]}{\sum k_i \tau_u[t] V_u[t]},
\label{eq:cot}
\end{equation}
\begin{equation}
R[t]= \sum_{\tau = t}^{t+H} \left( r[\tau] \right) + \min_{\tau = t}^{t+H}{ r[\tau]},
\label{eq:cotrwd}
\end{equation}
\noindent where $H$ denotes the horizon set as 14 timesteps, $m$ denotes the robot mass (4.7 kg), $g$ denotes the acceleration due to the earth gravity (9.81 m/s), $k_i$ denotes the mapping gain from the motor torque in \href{https://emanual.robotis.com/docs/en/dxl/x/xm430-w350/}{\url{https://emanual.robotis.com/docs/en/dxl/x/xm430-w350/}}, $\tau_u[t]$ denotes the torque of motor $u$, and $V_u[t]$ denotes the corresponding operating voltage. 

The experimental results are presented in terms of five aspects: (1) primitive locomotion learning on regular terrain, (2) general continual locomotion learning, (3) separation and incrementation of knowledge/behavior, (4) exploitation of similarity, and (5) interpretation and modification. While discussing the results obtained from full \frame{} in each aspect, the results are compared with different ablated versions to present the importance of such components and compared to the state-of-the-art methods that lack the equivalent mechanisms, as summarized in Table~\ref{tab:ablation}.

\begin{table}[!h]
	\centering
	\caption{Overview of the ablation study, where three main key components of \frame{}: the neural control, neurogenesis, and dual learning, are studied. \Checkmark indicates the component that are included during the ablation study, ``ablated'' indicate the components that are ablated during the ablation study, and - indicates the components that are excluded during the ablation study.}%
	\begingroup
	\def\y{{\footnotesize \Checkmark}}
	\def\ex{ablated}
	\def\n{-}
	\begin{tabular}{@{}c c c c c@{}}
		\toprule
		\multirow{2}{*}{\bd{Aspect}} & \multirow{2}{*}{\mc{\bd{Neural}\\\bf{Control}}}  & \multirow{2}{*}{\mc{\bd{Neuro}\\\bd{genesis}}}& \multicolumn{2}{c}{\bf{Dual Learning}}  \\ \cmidrule{4-5}		
		& & & \bd{Primary} & \bd{Supple.} \\
		\midrule
		Aspect~\ref{sec:exp_normal} & \ex & \n & \ex & \n \\ \midrule
		Aspect~\ref{sec:expgenerallearning} & \y & \y & \y & \y \\ \midrule
		Aspect~\ref{sec:expprimarylearning} & \y & \ex & \y & \y \\ \midrule
		Aspect~\ref{sec:expsupplementarylearning} & \y & \y & \y & \ex \\ \midrule
		Aspect~\ref{sec:expinterpretation} & \y & \y & \y & \y \\
		\bottomrule
	\end{tabular}
	\endgroup
	\label{tab:ablation}
\end{table}

\subsection{Primitive Locomotion Learning}
 \refstepcounter{rnum}
\label{sec:exp_normal}

\figexpnormal{} and a video at \videonormal{} reveal that the primitive locomotion learning on a regular flat terrain was achieved from scratch within the first 200 episodes ($\approx$ 10 mins) with final average walking speed of almost 10 cm/s after 10 repetitions.

At approximately 30 episodes, the robot started moving forward, as shown in the first row of \fignormalsnap. The robot began turning because it found that this could be a probabilistically simple strategy for receiving positive rewards. Merely 20 episodes after that, it demonstrated the capability of correcting its locomotion path and obtaining a faster forward speed, as shown in the second row of \fignormalsnap. By the 100$^{\text{th}}$ episode, the robot had developed a gait with a forward speed of approximately 5 cm/s on average, which is equivalent to the result obtained from a manually designed controller \citep{pbird_morfendocrine}, validated on the same robot. Finally, by the 200$^{\text{th}}$ episode, the average walking speed had reached almost 10 cm/s.
 
\begin{figure}[!h]
	\centering
	\includegraphics[width=0.85\linewidth]{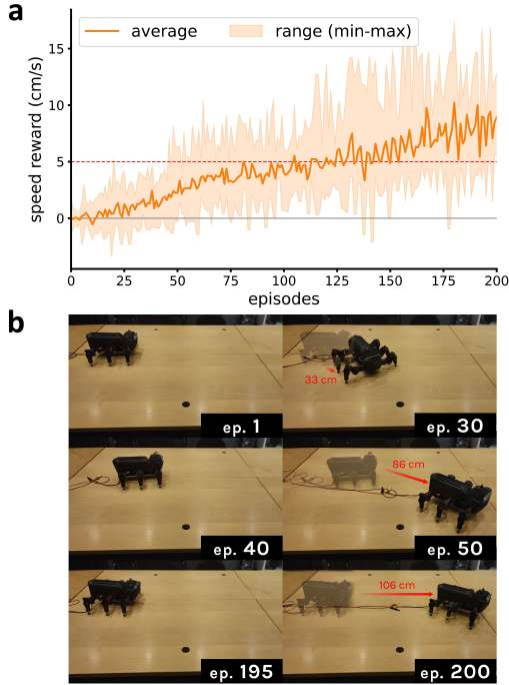}
	\caption{\sfig{a} Average episodic speed reward the physical robot locomotion learning on the \ttable{} (I) and \sfig{b} corresponding snapshots. The video is available at \videonormal.}
	\label{fig:expnormal}
\end{figure}

\begin{figure}[!h]
	\centering
	\includegraphics[width=0.8\linewidth]{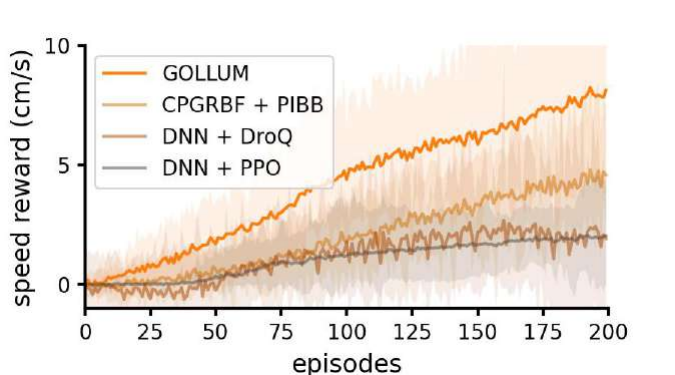}
	\caption{Average speed reward per gait cycle and its range (min--max), obtained from the simulated hexapod robot trained with different methods (\frame, CPGRBF + PIBB, DNN + DroQ, and DNN + PPO, see text for details). Note that the hyper-parameters of the methods (see the \sup) were obtained from grid search, performed around the values reported in their original works.}
	\label{fig:rewardcomparesim}
\end{figure}

To compared with the state-of-the-art methods, we conducted an extensive experiment. The locomotion learning of the simulated hexapod robot was evaluated on regular flat terrain using four different techniques, including \frame, a CPG-based technique (CPGRBF + PIBB) \citep{mathias_cpgrbf,mathias_nature}, an off-policy deep reinforcement learning technique (DNN + DroQ) \citep{droq,walkinthepark}, and an on-policy deep reinforcement learning technique (DNN + PPO) \citep{ppo}. The results, presented in Figure~\ref{fig:rewardcomparesim}, demonstrate \frame's advantage in learning speed. The robot using \frame{} achieved a speed of 5 cm/s after just 100 episodes, and nearly 10 cm/s by episode 200 (similar to Figure~\ref{fig:expnormal}). In contrast, CPGRBF + PIBB \citep{mathias_cpgrbf,mathias_freqeuncy} reached a final speed of only 5 cm/s, which is 50\% lower than \frame{} (p-value $<$ 0.05, t-test, n = 20). DNN + DroQ and DNN + PPO methods achieved even lower speeds, reaching only 2 cm/s per gait cycle, or 80\% lower than \frame{} (p-value $<$ 0.05, t-test, n = 20). Interestingly, the robots trained with DNN + DroQ and DNN + PPO exhibited unnatural, chaotic movements and increased gait frequency to achieve higher speeds, rather than adapting their locomotion patterns like those trained with \frame{} and CPGRBF + PIBB. A video of this experiment and comparison can be seen at \videosim.

Taken together, these results highlight \frame's sample efficiency. It achieves higher rewards and performance within the same number of learning episodes (or uses less learning time to reach equivalent reward levels) compared to the state-of-the-art methods, including CPG-based \citep{mathias_cpgrbf,mathias_nature} and deep reinforcement learning techniques \citep{droq,walkinthepark,ppo,dogrobot_massivelyparallel}, in the context of locomotion learning. This efficiency advantage makes \frame{} more practical for real world applications, as it requires significantly less learning time (on the order of minutes) to achieve comparable performance compared to previous methods, which typically require from hours to days in simulation \citetraining.

\subsection{General Continual Locomotion Learning}
 \refstepcounter{rnum}
\label{sec:expgenerallearning}

Extending from one environmental condition to multiple conditions, the robot demonstrated online continual locomotion learning on 4--6 new skills within an hours; each behavior/skill was learned under a similar timescale ($\approx$ 100--200 episodes or 10--20 mins). The new skills enabled the robot to walk on a deformable terrain \citep{kaise_scirobotics}, different slopes \citep{zumofirstgecko}, and a slope with motor dysfunction \citep{hexapod_legamp}. 

\begin{figure*}[!h]
	\centering
	\includegraphics[width=0.8\textwidth]{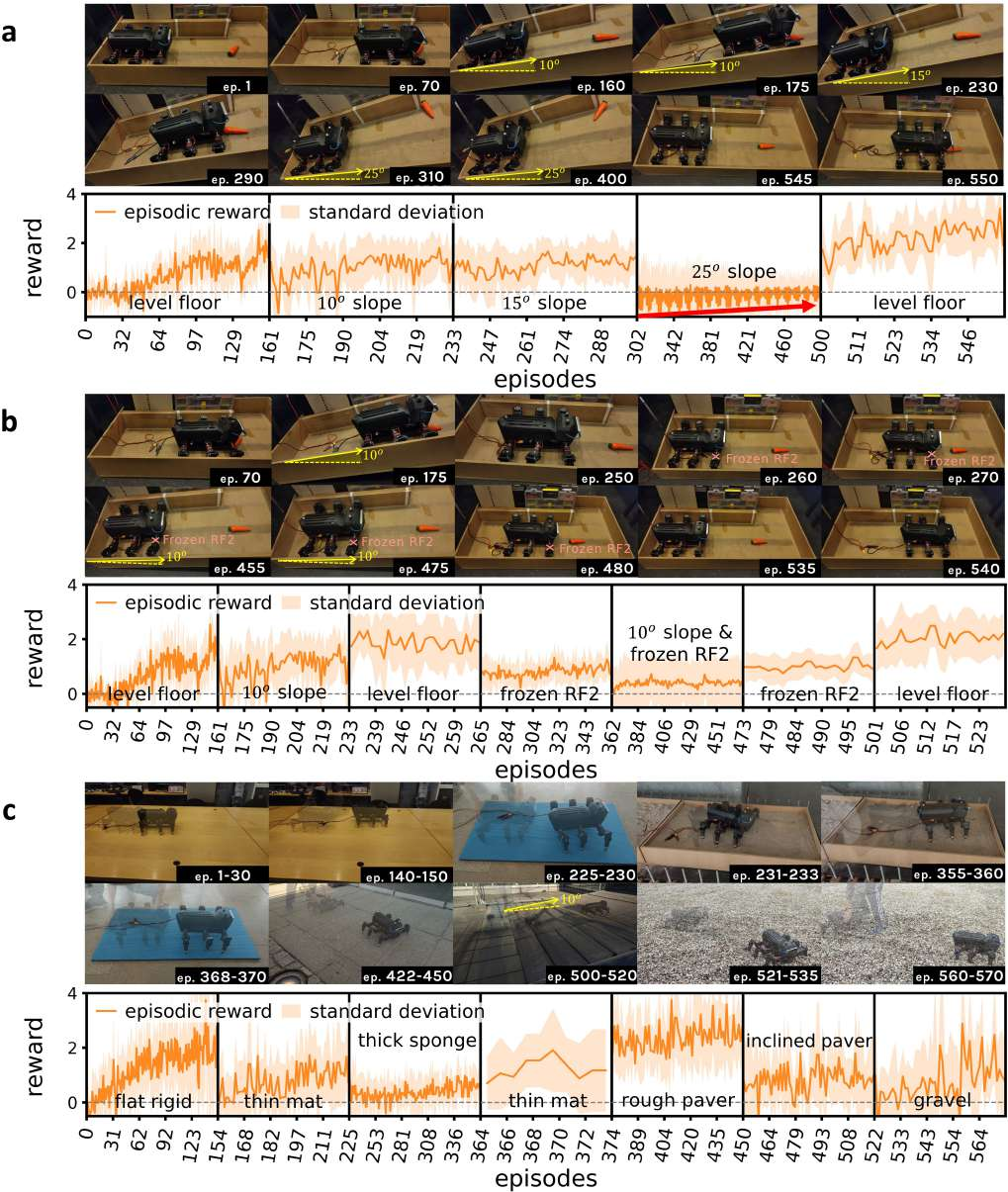}
	\caption{Snapshots and inverse cost of transport (COT)-based rewards (Eq.~\ref{eq:cotrwd}) obtained from a physical hexapod robot under online continual locomotion learning \sfig{a} \firstsequence{} (0\deg, 10\deg, 15\deg, and 25\deg), \sfig{b} \secondseqeunce{} (0\deg, 10\deg, 0\deg{} with RF2 dysfunction, and 10\deg{} with RF2 dysfunction), and \sfig{c} \thirdseqeunce{} (\ttable, \bluemat, \sponge, \rough, \roughslope, and \gravels). Details plots of the signals along with the learned foot trajectories are provided in the \sup, while the videos of the \exp s are available at \videoslope, \videobroken, and \videoterrain}
	\label{fig:expreward}
\end{figure*}

During the locomotion learning \firstsequence, shown in Figure~\ref{fig:expreward}\sfig{a}, the robot started on a confined level floor with no \review{previously learned} knowledge \review{(all output mapping weights are zero)}. It took merely 70 episodes to develop a proper locomotion pattern, receiving a reward of 0.5 (COT $\approx$ 190) and reaching the end of the platform, and it took 150 episodes to triple the reward up to 1.8 (COT $\approx$ 50). At approximately 160 and 230 episodes, the platform was inclined to 10\deg{} and 15\deg, respectively. This involved a difficulty in climbing up the slope at different angles; thus, the reward decreased to almost 1.0 (COT $\approx$ 90). To deal with these changes, the robot exploited previous knowledge to autonomously and continuously learn to find new locomotion skills. Based on this learning mechanism, it took merely 20 additional learning episodes ($\approx$ 2 mins) approximately to increase the reward to 1.2 (COT $\approx$ 80). After that, at approximately 300 episodes, the slope was further increased to 25\deg, which was beyond the robot capability. Interestingly, although the reward remained around 0.0, indicating that the robot did not move forward climbing, it learned to reduce the degree in sliding backward, as illustrated by the increase in the minimum reward and highlighted by the red arrow (Figure~\ref{fig:expreward}\sfig{a}). Finally, at approximately 500 episodes, the platform was returned to 0\deg. The robot could quickly recovered its locomotion to a regular gait for walking on the level floor. It later improved the locomotion and increased the reward from 1.8 (COT $\approx$ 50) to approximately 2.6 (COT $\approx$ 40) at 150 episodes.

During the locomotion learning \secondseqeunce, shown in Figure~\ref{fig:expreward}\sfig{b}, the robot also started on a confined level floor with no \review{previously learned knowledge (all output mapping weights are zero)}, before experiencing a 10\deg{} slope at approximately 160--230 episodes and returning to the level floor at approximately 230--260 episodes. However, after 260 episodes, the second motor of the right front leg (RF2, see \figoverview) was frozen (kept fixed), simulating the Dynamixel motor over-torque protection mechanism, where the RF leg obstructed the movement instead of contributing the locomotion, resulting in the drop of the reward from 1.8 (COT $\approx$ 50) to 0.8 (COT $\approx$ 130). Nevertheless, the robot could quickly deal with this incident and adapted its motion to increase the reward from 0.8 (COT $\approx$ 130) to 1.0 (COT $\approx$ 100) by the 350$^{\text{th}}$ episode (using $\approx$ 2 mins). At approximately 360 episodes, the platform was inclined again to 10\deg{}, simultaneously introducing two difficulties (i.e., slope and motor dysfunction) for learning, causing a reward reduction to 0.00. Interestingly, having learned the locomotion on slope and that with motor dysfunction, the robot took merely required 30 additional episodes to discover a proper locomotion pattern with a reward of 0.4 (COT $\approx$ 230) and slowly climbed up the slope. Finally, when the platform was returned to 0\deg{} at approximately 480 episodes and the RF2 motor was kept fixed at approximately 500 episodes, the robot achieved the same reward as that at approximately 360 episodes and 260 episodes, respectively.

During the locomotion learning \thirdseqeunce{} shown in Figure~\ref{fig:expreward}\sfig{c}, the robot started on a \ttable{} with no \review{previously learned} knowledge \review{(all output mapping weights are zero)}, where it began moving forward with a reward of 0.8 after 30 learning episodes ($\approx$ 3 mins) and achieved a reward of 2.0 (COT $\approx$ 49) after 150 episodes ($\approx$ 15 mins). After that, the robot was transferred to a soft \bluemat, where the reward dropped significantly to nearly 0.00 and then improved to 1.0 (COT $\approx$ 90) after 50 additional episodes ($\approx$ 6 mins). After 220 episodes, the robot was transferred to a \sponge{} terrain, where the robot's feet got stuck owing to the deformation of the terrain. However, it required approximately 100 episodes of learning to start developing a bouncing gait, bouncing on the \sponge{} to avoid getting stuck and receiving a reward of 0.6 (COT $\approx$ 170). After 360 episodes, the robot was transferred back to the \bluemat{} to demonstrate recalling a previously learned skill, and after 370 episodes, it was transferred to a \rough{} to learn a new skill. Interestingly, on the \rough, the robot found that the locomotion pattern for a \ttable{} could be used here given that it yielded a similar reward of approximately 2.0, which was then slightly increased to 2.5 (COT $\approx$ 40) in the following 80 episodes ($\approx$ 8 mins). After 450 episodes, the robot reached an \roughslope, where the reward dropped to 0.6 (COT $\approx$ 160), but it later increased to 0.8 (COT $\approx$ 120) after 60 episodes ($\approx$ 6 mins). Finally, the robot was transferred to a \gravels, where reward dropped to nearly 0.00.; however, it was able to find a new gait for which the reward was increased to 1.2 (COT $\approx$ 80) after 50 episodes ($\approx$ 6 mins).  

All these results demonstrate that, unlike several previous works \citep{hexapodlearning_fc_sixmodule,fourleg_cpg_multihead,plastic_matching,LILAC_latent_method,hexapod_blindlearning,MELA,mathias_cpgrbf,mathias_nature}, which often employ fixed networks after training, \frame{} can enable the robot to continuously and autonomously improve its locomotion to handle multiple (unseen) conditions less than 10 minutes per condition, even without prior simulation-based pretraining. The robot successfully adapted its locomotion patterns to walk up slopes near its hardware limits and cope with motor dysfunction within a confined testing platform. Moreover, the robot could also learn to handle real-world terrains, including highly deformable ones, such as \sponge s and \gravels s, which are difficult to accurately model. The results presented in the following sections aim to verify that \frame{} can also maintain previously learned locomotion patterns (i.e., \overfor) and efficiently utilize learned locomotion patterns to quickly find new patterns for new conditions (i.e., leveraging knowledge exploitation).

\subsection{Separation and Incrementation of Knowledge/Skills}
 \refstepcounter{rnum}
\label{sec:expprimarylearning}

\begin{figure*}[!h]
	\centering
	\includegraphics[width=0.9\textwidth]{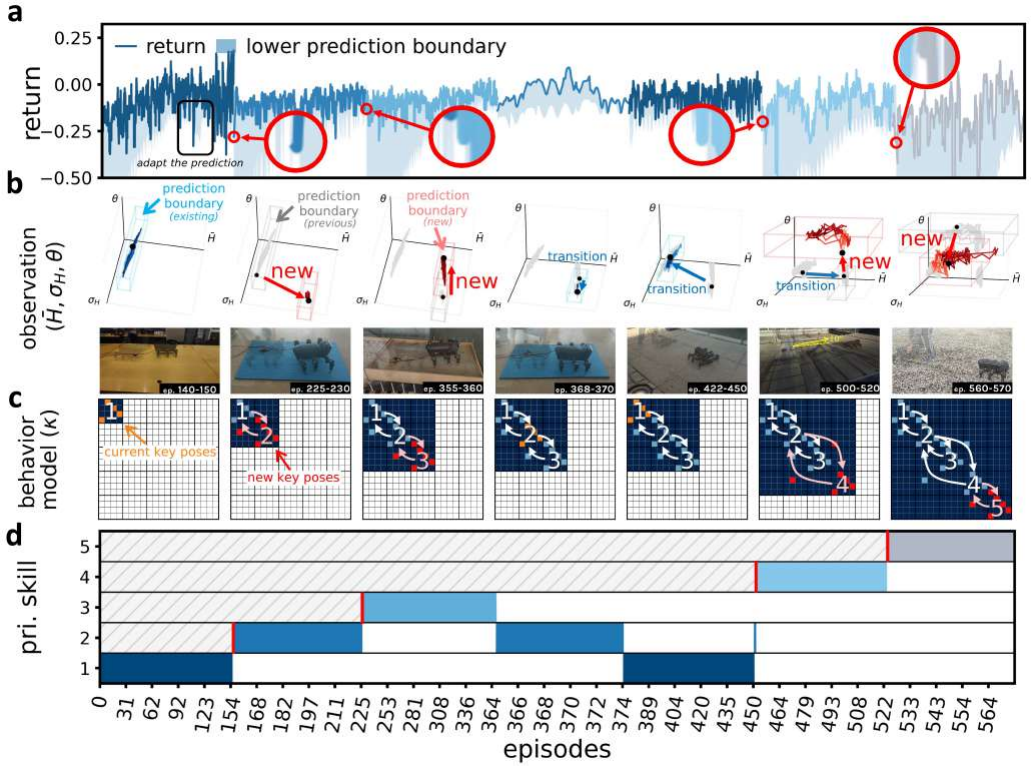}
	\caption{\sfig{a} Returns and lower prediction boundary obtained from locomotion learning \thirdseqeunce. \sfig{b} Trajectory of the sensory feedback in the observation space (hue mean $\bar{H}$, hue standard deviation $\sigma_H$, and body pitch $\theta$) presented along with its prediction boundary and snapshot of the terrains. \sfig{c} Robot behavior model extracted from and presented along with the transition matrix $\kappa$, i.e., the structure of hidden state connections, where the element at row $i$ and column $j$ ($\kappa_{ij}$) denotes the existence of the transition/positive connection from hidden state $i$ to $j$. \sfig{d} Activation of the primary skills. Note that, in this figure, the addition of new subnetworks is highlighted in red.}
	\label{fig:expprimarylearning}
\end{figure*}

Figure~\ref{fig:expprimarylearning} plots the return, observation, network structure, and subnetwork activities, recorded from continual locomotion learning \thirdseqeunce, to further demonstrates the separation of knowledge/skills and the incrementation process. Overall, as highlighted in red, the neurogenesis was triggered when the return dropped below the lower value prediction boundary, as shown in \figprimaryvalue{}, and the observation exceeded the previous observation prediction boundary, as shown in \figprimaryobservation. This created new subnetworks/neurons, as illustrated in \figprimarymatrix{}, and activated the learning of the corresponding subnetworks/skills, as illustrated in \figprimaryskill.

Between 1--150 episodes, the robot had only one untrained subnetwork, i.e., the set of four neurons indicated in orange in \figprimarymatrix, which was used and trained, as shown in \figprimaryskill. During this period, the robot also learned to predict the return and observation, and adapt their prediction boundaries  (i.e., uncertainty) to cover all the data points, as shown in Figures~\ref{fig:expprimarylearning}\sfig{a} and \sfig{b}. However, when the robot was transferred to the \bluemat{} at approximately 150 episodes, the first locomotion pattern/skill could no longer produce high return, causing the return falling below the lower value prediction boundary (i.e., shaded region in \figprimaryvalue), and the observation went over the previous observation prediction boundary (i.e., gray box in \figprimaryobservation). These two events together triggered the neurogenesis, which created the second set of neurons/subnetworks, as illustrated in red in \figprimarymatrix; moreover, their connection weights were also initialized with those of the previous set (refer to as \init{} mechanism) to facilitate the learning. Following that, the robot used and trained the second subnetwork/primary skill, as shown in \figprimaryskill. This process repeated when the robot was transferred to the \sponge{} terrain at approximately 220 episodes, creating the third subnetwork/primary skill. Note that, a significant reduction solely in the return did not trigger the neurogenesis; nevertheless, it expanded the value prediction boundary and prepared the robot for a sudden change in reward/return, as can be observed at approximately 100 episodes. 

When the robot was re-transferred to the \bluemat{} at approximately 360 episodes, it successfully recalled the second subnetwork learned previously using observation, as shown in \figprimaryobservation. Interestingly, when being transferred to the \rough, there was no significant change in both return and observation, as shown in \figprimaryvalue,\sfig{b}; as a result, the robot autonomously switched to the first primary skill (locomotion on \ttable), which exhibited the most similar observation patterns without creating any new subnetwork. This could also be considered a strategy for \expsim.

When the robot was on the \roughslope{} at approximately 450 episodes, the robot immediately switched from the first behavior/primary skill to the second one, given that the observation pattern received was more similar to the second condition (\bluemat) than the other. However, after trying the second primary skill, it found that the return was still below the low prediction boundary while the observation also exceeded the previous observation previous prediction boundary (i.e., gray box in \figprimaryobservation). As a result, the fourth subnetwork was created and connected to the previously active one (i.e., the second subnetwork, locomotion learning on \bluemat), as illustrated in \figprimarymatrix. Interestingly, leveraging the \init{} mechanism, the new subnetwork was not trained entirely from scratch; it was initialized with the connection weights of the previously active skill (i.e., locomotion on the \bluemat), which was the most similar ones in the observation space, and underwent almost instantaneously transition based on the activity of the neurons, as demonstrated in \figprimaryskill. Starting from such activity, the robot kept using and refining the fourth primary skill until it encountered the \gravels, where the fifth subnetwork was autonomously created following a similar processes. Note that, this behavior model, i.e., the organization of behaviors (refereed to \behav), can be observed from the connection matrix (i.e., $\kappa$, which represents also the connections between \ZPG{} neurons), as illustrated in \figprimarymatrix. The visualizations of locomotion learning \firstsequence{} and \secondseqeunce{} are available in the \sup.

\begin{figure}[!h]
	\centering
	\includegraphics[width=0.7\linewidth]{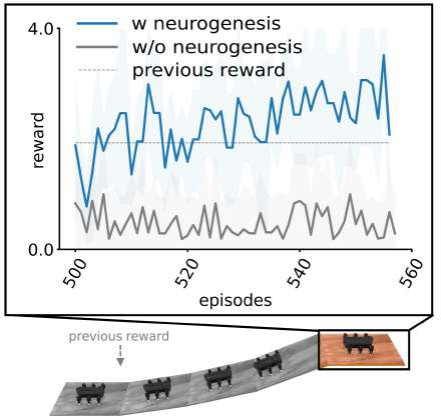}
	\caption{Comparison of the rewards from locomotion learning \firstsequence{} (blue line) with and (gray line) without neurogenesis.}
	\label{fig:expprimarylearning1}
\end{figure}

\begin{figure}[!h]
	\centering
	\includegraphics[width=0.9\linewidth]{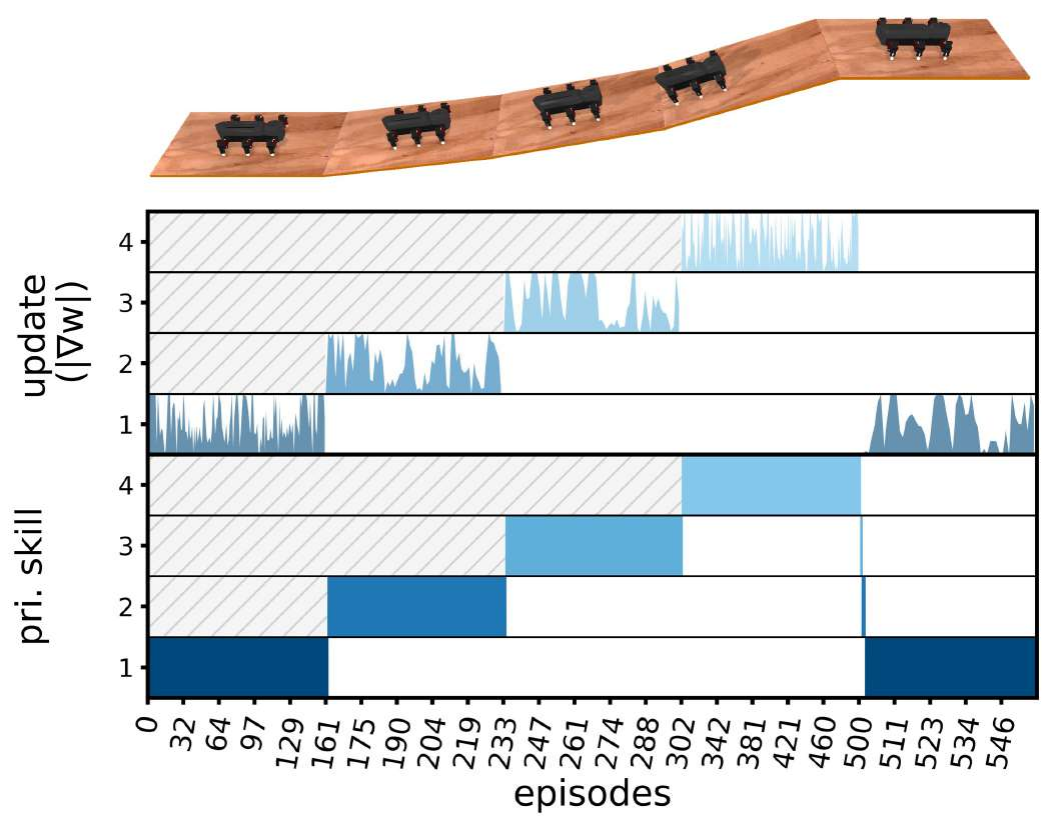}
	\caption{The normalized magnitude of the weight updates ($|\nabla w|$) computed from the learning rule during the learning, presented alongside the activation of four subnetworks (i.e., the activation/usage of the primary skills) and the training sequence. Given that the connection weights encode the knowledge of the behaviors/skills, non-zero weight updates indicate changes in the corresponding behavior/skill, while zero weight updates indicate no change. In this example, MORF walked from a level floor up a slope with varying angles and then returned to the level floor. Note that the blue color with different shades represents the activation of different primary skills and the corresponding weight updates, while the gray diagonal line pattern indicates that the primary skills have not yet been learned.}
	\label{fig:noforget}
\end{figure}

Figure~\ref{fig:expprimarylearning1} shows that the robot suffers from catastrophic forgetting if it performs locomotion learning \firstsequence{} without any knowledge separation mechanism. With the proposed neurogenesis (blue line), the locomotion skill on 0\deg{} learned previously was successfully recalled, and the robot received a reward of approximately 1.8, which was later increased to approximately 2.6 as a result of continual learning. Figure~\ref{fig:noforget} further reveals that, when using neurogenesis, \frame{} did not update the connection weights of inactive subnetworks. As a result, those inactive behaviors/skills remained unchanged; thereby preventing catastrophic forgetting. By contrast, without the neurogenesis (gray line), the locomotion on 0\deg{} learned previously was interfered by the locomotion on 25\deg, which focused on reducing sliding backward instead of moving forward. This is because all the connection weights changed throughout the training sequence when only one subnetwork was used. Thus, the reward remained at approximately 0.4 throughout the training, or rather merely 20\% of the previous value (p-value $<$ 0.05, paired t-test, n=20). Therefore, the previous approaches that do not have an \overfor{} mechanism \citep{hexapodlearning_fc_sixmodule,fourleg_cpg_multihead,plastic_matching,LILAC_latent_method,hexapod_blindlearning,MELA,mathias_cpgrbf,mathias_nature} will experience catastrophic forgetting when they are trained continuously without being frozen.

\begin{figure}[!h]
	\centering
	\includegraphics[width=0.7\linewidth]{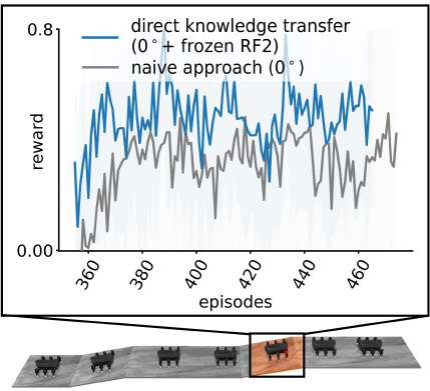}
	\caption{Comparison of the rewards from locomotion learning \secondseqeunce{} when the fourth subnetwork/skill was initialized (blue line) with the locomotion on 0\deg{} slope with frozen RF2 (\init) and (gray line) with regular locomotion on 0\deg{} (naive approach).}
	\label{fig:expprimarylearning2}
\end{figure}

Figure~\ref{fig:expprimarylearning2} also shows that, after adding a new subnetwork, the skill initialization could affect performance at the very first few episodes. Using \frame{} with the \init{} mechanism (blue line) automatically selected the previously learned knowledge according to the most similar skill on observation-based behavior transition. In this case, the robot chose the locomotion pattern on 0\deg{} floor with frozen RF2 as the initialization of the fourth behavior/skill. As a result, the reward started at 0.36 around 360 episodes before increasing to around 0.60 after 460 episodes. However, when simply using the regular locomotion pattern (gray line), the locomotion on 0\deg{} floor was used as the initialization (referred to as the naive approach); as a result, the reward started around 0.0, which was significantly lower than that with the \init{} (p-value $<$ 0.05, paired t-test, n = 20). This indicates that initializing the new behavior/skill based on observation similarity using \init{} can result in a higher reward than simply initializing new behaviors/skills with the locomotion on regular flat floor. Therefore, the previous approaches that do not include this autonomous \init{} mechanism \citep{piggyback,HlifeRL_optionbased,KeepLearning} cannot autonomously obtain the head start performance. However, after few learning episodes, the rewards obtained from both testing conditions increased to similar levels (p-value $>$ 0.05, paired t-test, n=20) due to the supplementary learning (Figure~\ref{fig:expprimarylearning2}). Next, in the following aspect, the exploitation of similarity between different skills/environments using supplementary learning is further investigated.

\begin{figure*} 
	\centering
	\includegraphics[width=0.9\textwidth]{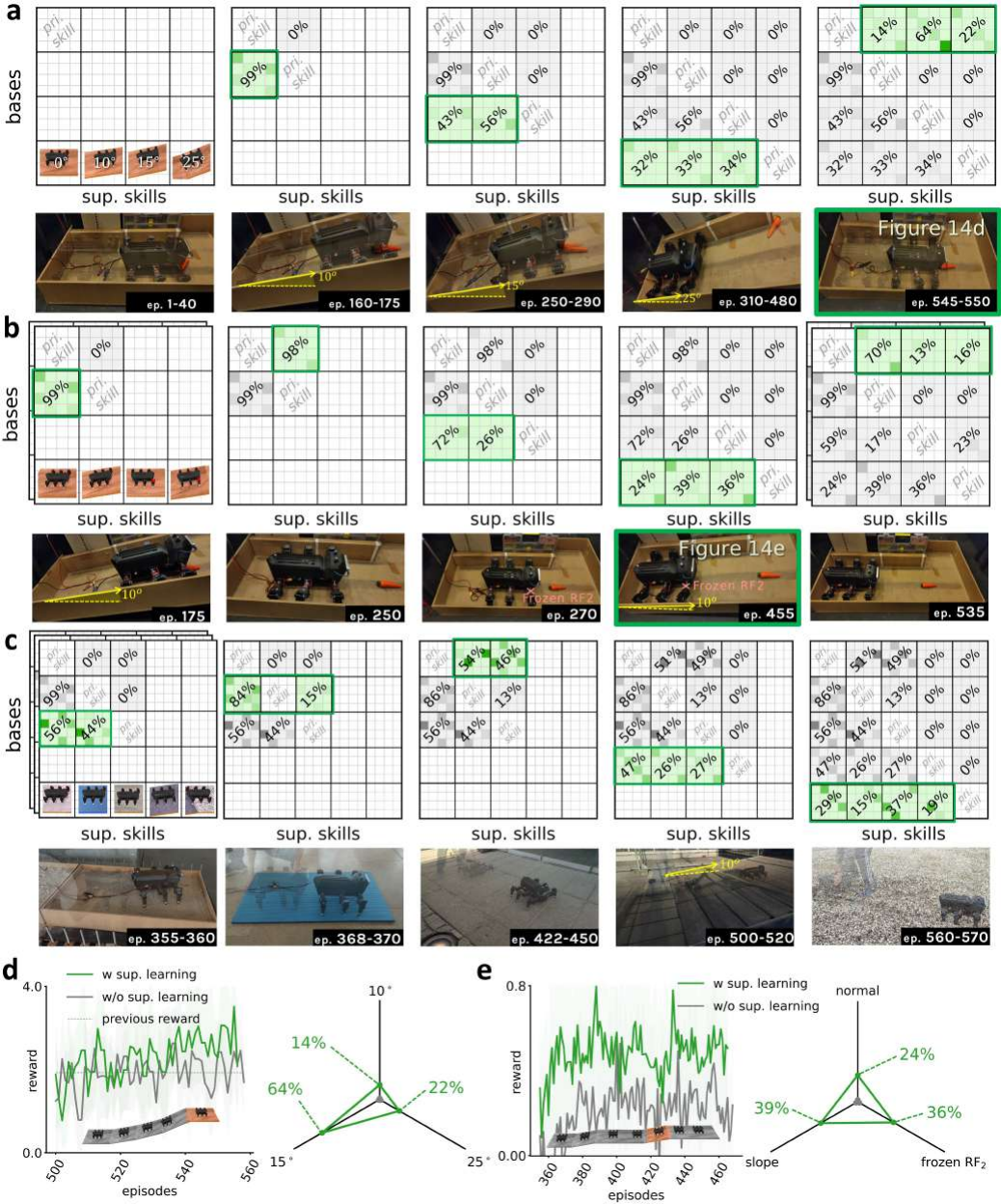}
	\caption{Snapshots and connection weight matrices for exploiting task similarity obtained from the locomotion learning \sfig{a} \firstsequence, \sfig{b} \secondseqeunce, and \sfig{c} \thirdseqeunce. The rows highlighted in green indicate the supplementary skill contribution percentages (i.e., the current exploitation ratio of all skills) under different environments. \sfig{d,e} Comparison of the rewards and supplementary skill contribution percentages obtained from the locomotion learning \firstsequence{} and \secondseqeunce{} (green line) with and (gray line) without supplementary learning.}
	\label{fig:expsuplementarylearning}
\end{figure*}

\subsection{Exploitation of Similarity}
 \refstepcounter{rnum}
\label{sec:expsupplementarylearning}

While the robot primarily uses and learns intra-subnetwork connections from high-level patterns (PM) to motor commands (M), it also learns the supporting inter-subnetwork connections from internal states/bases (B) to action patterns encoded in the premotor neurons (PM), as depicted in \figoverview, to supplement the combination of all acquired skills under adaptive combination percentages. Accordingly, the contribution percentages can be obtained by visualizing such connection weights, as shown in Figures~\ref{fig:expsuplementarylearning}\sfig{a}--\ref{fig:expsuplementarylearning}\sfig{c}, where the weight matrix represents the magnitude of each supplementary skill (i.e., each column) according to each internal state (i.e., each row). In general, the robot learned and adapted the supplementary contributions that corresponded to the active bases, as highlighted in green, while maintaining those contribution that corresponded to inactive bases to prevent forgetting as depicted in gray.

As a first example, Figure~\ref{fig:expsuplementarylearning}\sfig{a} presents the supplementary contribution learned during locomotion learning \firstsequence, where the robot exploited the locomotion skills learned on lower slopes for supporting the learning on steeper slopes, and vice versa. Initially, the robot had acquired only the first skill from the confined level platform; thus, there was no exploitation of similarity in this stage. Later, when the robot was on a 10\deg{} slope, \frame{} automatically created and trained the second skills and supplemented the first skill (i.e., locomotion on 0\deg) to ease the learning. Similarly, on 15\deg{} and 25\deg{} slopes, \frame{} created and trained the third and fourth skills, respectively, and supplemented other locomotion skills with a slightly greater contribution offered from the locomotion skill on higher slopes. Interestingly, the supplementary learning proved crucial when the robot returned to the level platform after experiencing a 25\deg{} slope, which was beyond the robot capability (see Figure~\ref{fig:expsuplementarylearning}\sfig{d}). Without the supplementary learning, the robot autonomously switched to the first locomotion skill (0\deg) and was unable to access other skills, resulting in a consistent reward of approximately 1.8 (p-value $>$ 0.05, paired t-test, n=20), as depicted by the gray line. In contrast, with supplementary learning enabled, the robot effectively supplemented 64\% of the supplementary skill from the locomotion skill on a 15\deg{} slope (i.e., the maximum slope afforded by the hardware) and 22\% from the 25\deg{} slope, reducing backward sliding without human intervention. This led to the increase of the reward from 1.8 to 2.6, representing a 40\% improvement (p-value $<$ 0.05, paired t-test, n=20), within 60 additional episodes. Previous approaches lacking a mechanism for autonomously exploiting similar knowledge/skills \citep{hexapod_legmap,BaysianIncremental,KeepLearning} were unable to exploit this advantage.

As a second example, Figure~\ref{fig:expsuplementarylearning}\sfig{b} presents the supplementary contribution learned during locomotion learning \secondseqeunce, where the robot learned to exploit the locomotion skill trained on slope and that trained with motor dysfunction to ease the locomotion learning both on slope and with the motor dysfunction. Unlike the previous example, after the robot was transferred from the slope back to the level platform, it autonomously recalled the locomotion skill for the level platform while supplementing the locomotion skill for slope to facilitate the learning. Subsequently, after the RF2 motor was frozen, the robot autonomously learned to exploit the locomotion skill for the level platform at 72\% contribution given that it was on 0\deg{} slope, and supplemented the locomotion skill on slope at 26\% contribution. Interestingly, later on, when the RF2 motor was frozen on the slope, the robot used 39\% of the supplementary skill contribution from the locomotion skill on slope to deal with the inclined platform and 36\% of that contribution with the frozen RF2 motor to deal with the motor dysfunction. This resulted in an increase of the reward from nearly 0.0 to approximately 0.5 in merely 20 episodes, as shown by the green line in Figure~\ref{fig:expsuplementarylearning}\sfig{e}. However, when the supplementary learning was disabled during this stage, the robot could neither access nor utilize the locomotion skills on slope and with frozen RF2 motor, causing a the reward to increase to 0.2 under a similar learning time, or rather 60\% less (p-value $<$ 0.05, paired t-test, n = 20), as shown by the gray line in Figure~\ref{fig:expsuplementarylearning}\sfig{e}. Finally, after the RF2 motor was simulated being repaired and the robot returned to the level platform, the robot learned to supplement a majority 70\% of the locomotion on slope while using minor contributions from those related to motor dysfunction.

As a third example, Figure~\ref{fig:expsuplementarylearning}\sfig{c} presents the supplementary contribution learned during the locomotion learning \thirdseqeunce, where the robot exploited and combined the locomotion skills learned under different terrains. Initially, the robot acquired the first and second locomotion skills for the \ttable{} and \bluemat{}. On the \sponge, the robot learned to utilize 56\% of the supplementary contribution from the locomotion on the \ttable{} plus 44\% from that on the \bluemat{} to prevent its legs getting stuck in the thick deformable (soft) terrain. Next, after returning to the \bluemat, the robot learned to incorporate the newly acquired skill as the contribution rose from 0\% to 15\%. After that, on the \rough, it found that the locomotion on the \ttable{} could be reused; additionally, it supplemented the locomotion skills on the \bluemat{} and \sponge{} to ease the learning, contributing to a slight increase of the reward, as shown in Figure~\ref{fig:expreward}\sfig{c}. Interestingly, on the \roughslope, the robot autonomously learned to supplement a majority of 47\% from the locomotion on the \ttable/\rough{} given that the \roughslope{} was similar to the \rough{}, differing only in the inclination. Finally, on the \gravels, it used 37\% of the contribution from the locomotion on the \sponge{} to deal with the deformable nature of the \gravels{}, combined with 29\% of the contribution from the locomotion on the \ttable{}/\rough{} to deal with this rough terrain.

\subsection{Interpretation and Modification}
\refstepcounter{rnum}
\label{sec:expinterpretation}

To present and validate the interpretability of \frame, this study employs both empirical demonstrations and quantitative comparison.

\begin{figure*}[!h] 
	\centering
	\includegraphics[width=0.98\linewidth]{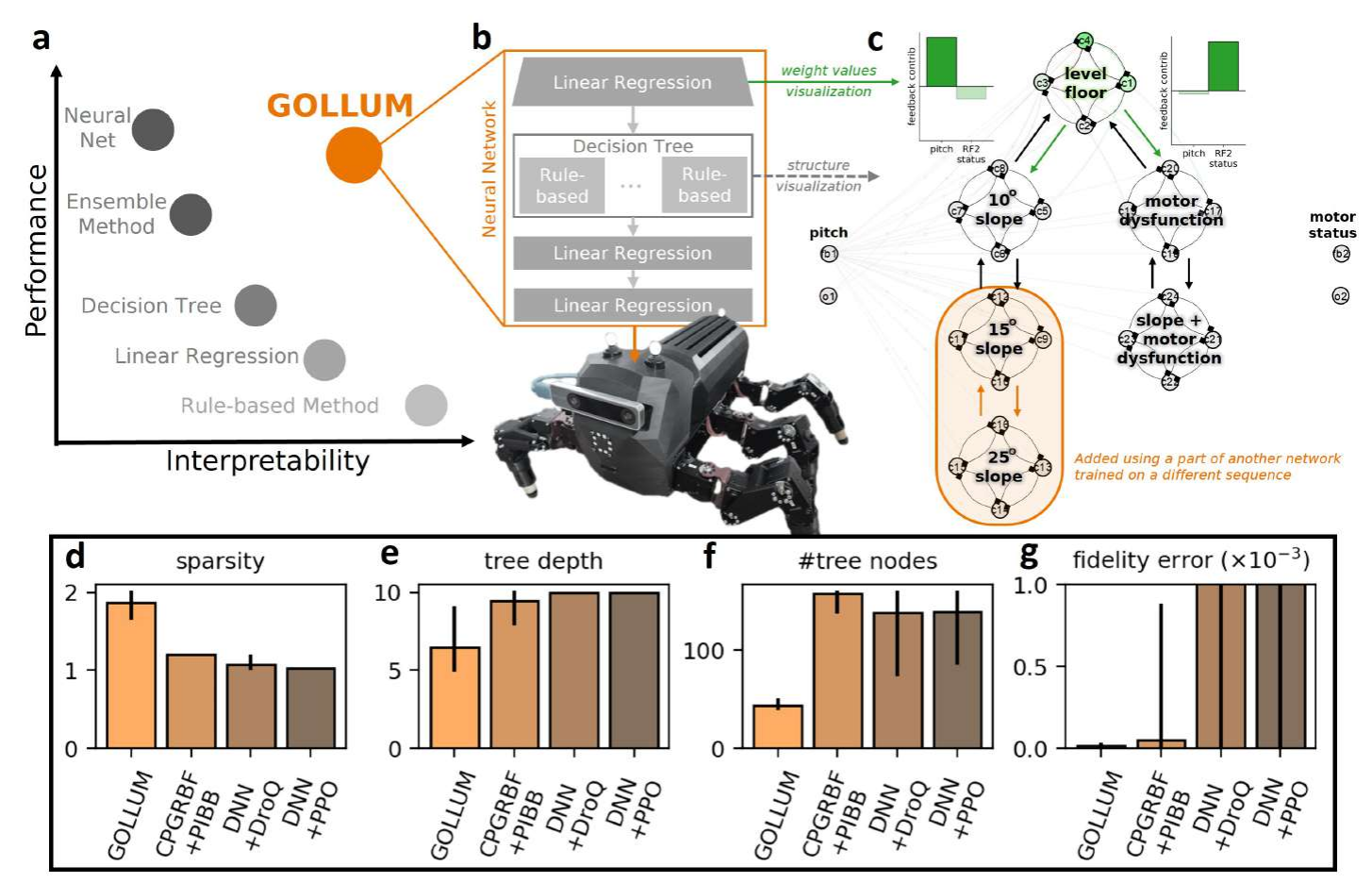}
	\caption{\sfig{a} Trade-off between interpretability and performance in various machine learning methods \citep{darpaxai}, including \frame, which has a higher interpretability and performance (see Figure~\ref{fig:rewardcomparesim}). \reviewfinal{This figure is a conceptual representation that broadly illustrates the overall trend rather than precise, scale-accurate measurements of interpretability and performance.} \sfig{b} \frame{} represented as the combination of multiple interpretable models: a linear regression for sensory preprocessing, a decision tree for internal state generation, a sparse linear regression for sharing different action patterns trained with supplementary learning, and another sparse linear regression for output mapping trained with primary learning. \sfig{c} Behavior model extracted from an interpretable neural control, specifically the structure of the sequential central pattern generator layer. The interpretable neural control is trained on a level floor and 10\deg{} slope with possible motor dysfunction before adding a portion of another interpretable neural control trained on multiple slopes, demonstrating modifiability. The extracted behavior model also includes feedback contributions for two behavior transitions (i.e., from the locomotion on level floor to the locomotion on a 10\deg{} slope and the locomotion with motor dysfunction), which are visualized directly from the weight values in the first sensory preprocessing layer. \sfig{d--g} Quantitative interpretation evaluation metrics, including compactness (the sparsity of the neural networks and the depth and the number of nodes of the post-hoc decision tree-based explanations) and completeness (the fidelity of the post-hoc decision tree-based explanations), presented along with the range (min--max). These metrics are compared across \frame, CPGRBF trained with PIBB \citep{mathias_cpgrbf}, DNN trained with DroQ \citep{droq,walkinthepark}, and DNN trained with PPO \citep{ppo,dogrobot_massivelyparallel}.}
	\label{fig:interpretation}
\end{figure*}

To empirically demonstrate the interpretability and its benefits, three characteristics of \frame: decomposability, transparency, and simulability \citep{taxonomiesXAI,review_mythos,reviewInterpretableRL}, are presented. Firstly, decomposability is achieved as \frame{} is built from different interpretable modular layers combined with column-wise subnetworks, making it a white-box model. As shown in \figoverview{} and \videodecompose, each component in an ``interpretation coordinate'' serves a specific function for a certain action, with each network parameter having a distinct function as summarized in the \sup. Secondly, transparency is reflected in how learning resembles training multiple stacked linear regressions with sparse inputs, as illustrated in Figures~\ref{fig:interpretation}\sfig{a,b}. The weights for behavior classification directly reflect the contributions of sensory feedback signals for behavior transitions (Figure~\ref{fig:interpretation}\sfig{c}). Additionally, the learned weights for output and pattern mapping are adjusted towards previously tried leg configurations/actions and previously accessed patterns/skills with high rewards (Figure~\ref{fig:learningplot} and Eq.~\ref{eq:rpg}). Lastly, simulability is demonstrated by the ability to convert a trained neural control network into a behavior model (Figure~\ref{fig:interpretation}\sfig{c} and \videobehavmodel), revealing the organization of complex behaviors and their transitions. By possessing these three key interpretation characteristics, \frame{} allows for understanding of how the robot adapts to different conditions. It achieves this by combining neural control networks trained under various conditions (Figure~\ref{fig:interpretation}\sfig{c} and \videobehavmodel) and adjusting network parameters, e.g., increasing the locomotion frequency parameter (increasing $\tau_i$, \videofreq), to achieve desired behaviors without the need for additional training.

To quantitatively assess interpretability, four XAI evaluation metrics were used, following \cite{evaluationXAI2}: sparsity (i.e., the inverse of the active neuron ratio (activation $>$ $10^{-3}$)), the depth of the post-hoc decision tree-based explanation, the number of nodes in the post-hoc decision tree-based explanation, and fidelity error (i.e., the mean square error between the outputs of the neural networks and the outputs of the post-hoc decision tree-based explanation). Figures~\ref{fig:interpretation}\sfig{d--g} illustrated the comparison between our technique, i.e., \frame, and three different state-of-the-art techniques, i.e., CPG-based control (CPGRBF trained with PIBB, \cite{mathias_cpgrbf}), off-policy deep reinforcement learning (DNN trained with DroQ, \cite{droq,walkinthepark}), and on-policy deep reinforcement learning (DNN trained with PPO, \cite{ppo,dogrobot_massivelyparallel}). Figure~\ref{fig:interpretation}\sfig{d} demonstrates that, due to merely half of the neurons being active at each timestep, the neural activities of \frame{} are 67--84\% significantly more sparse than the others (p-value $<$ 0.05, t-test, n = 20). This sparsity measurement reflects the simplicity/compactness of interpretation of the actual model without any further simplification. 

In addition to this, Figures~\ref{fig:interpretation}\sfig{e,f} reveal that the depth and the number of nodes of the decision tree-based explanation obtained from \frame{} are receptively 30\% and 70\% less than those from the others (p-value $<$ 0.05, t-test, n = 20). These measurements indicate the simplicity/compactness of the simplified post-hoc explanation obtained from each technique. Figure~\ref{fig:interpretation}\sfig{g} further shows that, even using less depth and fewer nodes in explaining \frame{}, the error is significantly less than those from the DNN trained with DroQ and DNN trained with PPO (p-value $<$ 0.05, t-test, n = 20).

\section{Discussion}
\refstepcounter{snum}
\label{sec:discussion}

\subsection{Life-long Locomotion Learning Research Aspect}

This study proposes a life-long locomotion learning framework, called \frame, for robot locomotion intelligence. It also demonstrates that interpretability (white-box machine learning) could be utilized to deal with four key challenges of locomotion learning.

\subsubsection{Sample Efficiency Challenge}

The first challenge is the extensive learning time typically reported, ranging from one hour to several days for locomotion learning on a flat terrain \citetraining. To this end, the \frame{} framework employs separately trained neural column structures representing actions (i.e., robot configurations: sets of all robot joint positions). This approach fast flat terrain locomotion learning on a physical robot, achieved in 10--20 minutes or 100--200 episodes (1 episode $\approx$ 1 gait cycle). Even in other conditions (such as on slopes, deformable terrains, and with a broken motor), the robot can autonomously learn the locomotion and achieve significant improvements under a similar timescale. The comparison between \frame{} and previous works in terms of learning time is summarized in the first four columns of Table~\ref{tab:compare}. Notably, compared to most existing approaches (in particular, 95 \% of the methods mentioned in Table~\ref{tab:compare}), which often rely on simulation and struggled to bridge the reality gap, \frame{} demonstrates locomotion learning in the real world using a significantly shorter timescale, or rather over 60\% reduction time. This fast learning time is achieved through the use of less correlated basis signals to ease collective learning \citep{RLbook}. Consequently, no human predefined assumption/constraints, such as indirect encoding \citep{mathias_cpgrbf,mathias_nature} or gait parameterization \citep{fourleg_cpg_multihead,hexapodlearning_cpg_spikingandforce3leg}, are required to reduce the optimization space. Thus, \frame{} offers a higher flexibility in locomotion learning than those methods \citep{mathias_cpgrbf,mathias_nature,fourleg_cpg_multihead,hexapodlearning_cpg_spikingandforce3leg}. Besides, \frame{} requires less sample estimate (i.e., merely 560 timesteps compared to 80,000 -- 1 million timesteps in other works \citep{dogrobot_teacherstudent_original,dogrobot_teacherstudent,MELA}) and enables the learning of multiple behaviors using a single simple reward term, i.e., forward speed or inverse cost of transport. This eliminates the need for biased human knowledge guidance and avoids the intensive selection of hyperparameters.

\begin{sidewaystable*}
	\caption{Comparison of different state-of-the-art locomotion learning methods in terms of continual locomotion learning.}\label{tab:compare}
	\centering
	
	\setlength{\dashlinedash}{0.8pt}
	\setlength{\dashlinegap}{1.5pt}
	\begin{tabular*}{\textheight}{@{\extracolsep\fill}lccccccc}
		\toprule
		
		\mc{\mrow{4}{\bd{method}}} & \multicolumn{3}{c}{\bd{locomotion learning (flat terrain)}} & \mcol{3}{\bd{continual locomotion learning ($>$ 1 skills)}} & \mrow{4}{\mc{\bd{human} \\ \bf{intervention}}} \\ \cmidrule{2-4} \cmidrule{5-7}
		
		& \mrow{2.5}{\mc{\bd{trained}\\\bd{robot}}} & \mrow{2.5}{\mc{\bf{rollout}\\\bf{(eps)}}}    & \mrow{2.5}{\mc{\bd{training}\\\bf{(eps)}}}  &  \mrow{2.5}{\mc{\bd{prevent}\\\bf{forgetting}}} & \mcol{2}{\bf{exploitation of similarity}} \\ \cmidrule{6-7}
		& & & & &  \bf{direct transfer} & \bf{learning} &\\
		\midrule 
		\makecell[l]{\cite{hexapodlearning_fc_sixmodule}} & \mc{\rd{sim}\\\rd{hexapod}} & \na &\mc{2.3k}  & \mcol{3}{\mrow{16}{\mc{\no\\\rd{(frozen after training)}}}} & \pretrain  \\  \cdashline{1-4} \cdashline{8-8}
		\cite{fourleg_cpg_multihead}  & \mc{\rd{sim}\\\rd{quadruped}} & \mc{64\\\gry{(128m steps)}} &\mc{100\\\gry{(6 hrs)}}  & & & & \pretrain  \\  \cdashline{1-4} \cdashline{8-8}
		\cite{plastic_matching} & \mc{\rd{sim}\\\rd{quadruped}} & 15 & \mc{150\\\gry{(1 hr)}} & & & & \pretrain \\  \cdashline{1-4} \cdashline{8-8}
		\cite{LILAC_latent_method} & \mc{\rd{sim}\\\rd{robots}} & yes & \mc{1.5k--3k\\\gry{(1--1.5 hrs)}} & & & & \pretrain  \\ \cdashline{1-4} \cdashline{8-8}
		\cite{hexapod_blindlearning} & \mc{\rd{sim}\\\rd{hexapod}} & 40  & \mc{10k--20k\\\gry{(2.5--25 hrs)}} & & & & \pretrain   \\ \cdashline{1-4} \cdashline{8-8}
		\cite{MELA} & \mc{\rd{sim}\\\rd{quadruped}} & \mc{20\\\gry{(1m steps)}}  & \mc{50--100\\\gry{(3--5 hrs)}} & & & & \pretrain    \\  \cdashline{1-4} \cdashline{8-8}
		\makecell[l]{\cite{mathias_nature}\\\cite{mathias_cpgrbf}} &\mc{\rd{sim}\\\rd{hexapod}} & \mc{8\\\gry{(2.8k steps)}} & \mc{80--240\\\gry{(28 mins)}} & & & & \rd{\mc{\pretrain{} \&\\encoding rule}}   \\ \cdashline{1-4} \cdashline{8-8}
		\makecell[l]{others$^\dag$} & \mc{\rd{sim}\\\rd{quadruped}} & \mc{{\footnotesize up to}\\4096} &\mc{ 0.4--0.8m\\\gry{(1--22 days)}} & & & & \pretrain  \\ 
		
		\midrule
		\cite{piggyback} & \mc{\rd{sim}\\\rd{quadruped}} & 4096 & 2m   &  \yes & \no & \yes & \pretrain \\ \cdashline{1-8}
		\cite{hexapod_legmap} & \mc{\rd{sim}\\\rd{hexapod}} & \na & \mc{40m \\\gry{(2 weeks)}}  & \yes & \yes & \no & \rd{\mc{\pretrain{} \& \\ collect options}}  \\ \cdashline{1-8}
		\cite{BaysianIncremental} &\mc{\rd{sim}\\\rd{other type}} & 16 & 300   & \yes &  \yes & \no & \pretrain{}  \\ \cdashline{1-8}
		\cite{KeepLearning} & \mc{\rd{sim}\\\rd{quadruped}}  & yes & \mc{5k--12k \\\gry{(2 hrs)}}  &  \yes &  \mc{\yes\\\gry{(task context)}} & \no & \rd{\mc{\pretrain{} \&\\ task context}} \\ \cdashline{1-8}
		
		\cite{HlifeRL_optionbased} & \mc{\rd{sim}\\\rd{walker}} & 64 & \mc{400\\\gry{(200 mins)}} & \yes &  \mc{\yes\\\gry{(task context)}} & \yes & \rd{\mc{\pretrain{} \&\\task context}}   \\  \cdashline{1-8}
		\cite{robotskillgraph} & \mc{\rd{sim}\\\rd{quadruped}} & 6144 & \na & \yes &  \yes & \yes & \rd{\mc{\pretrain{} \& collect \\ fundamental skills}}   \\ 
		\midrule
		\frame{} & \mc{physical\\hexapod} & \mc{8\\\gry{(560 steps)}} & \mc{100--200 \\\gry{(10 mins)}} & \yes & \yes & \yes  &  not required \\
		\bottomrule
	\end{tabular*}
	\raggedright
	\footnotesize{$^\dag$ \cite{dogrobot_massivelyparallel,dreamwaq,dogrobot_teacherstudent_original,dogrobot_teacherstudent}}
\end{sidewaystable*}

\subsubsection{Overcoming Catastrophic Forgetting \& Exploitation of Task Similarity Challenges}

The second and third challenges are jointly known as the dilemma of \say{\overfor-\expsim} \citep{lifelong_hamburg,bio_liftlonglearning}, which prevents robots from efficient learning and maintaining of multiple behaviors/skills throughout their life time. To deal with this, \frame{} employs a \duallearning. The first learning layer is called primary learning. It separately updates each primary skill encoded within each column/ring subnetwork to maintain the condition-specific skill. To exploit the other learned skills and facilitate learning, the second learning layer, called supplementary learning, updates the contribution shared between different skills/subnetworks. With this \duallearning, \frame{} successfully demonstrates the ability to maintain existing skills for \overfor{} and the ability to accelerate/improve the performance through the \expsim{} with other skills. 

As summarized in the last three columns of Table~\ref{tab:compare}, 12 out of the 18 presented locomotion learning methods (70\%) cannot perform continual learning given that the controller remains frozen to prevent catastrophic forgetting \citep{dogrobot_teacherstudent_original,MELA,hexapod_blindlearning,mathias_nature}. For examples, two-level policy \citep{hexapod_blindlearning}, MELA \citep{MELA}, and VMNC \citep{mathias_nature}, which adopt a straightforward mixture of experts, do not update the controller after training on a predefined set of environments. From 12 methods, six methods are capable of continual locomotion learning without forgetting. One of those, Piggyback \citep{piggyback}, requires some training before the skills encoded in share parameters are exploited because of the non-interpretable parameter structure of deep fully connected neural networks. Other five methods initialize new skills with the learned skills that have highest observation similarity \citep{BaysianIncremental,robotskillgraph}, have high value/reward after trial-and-error \citep{hexapod_legmap}, have been predefined by a given task-context \citep{HlifeRL_optionbased,KeepLearning}, or have been identified autonomously using observation and value/reward predictions (\frame). From those six methods, only three of them (HlifeRL \citep{HlifeRL_optionbased}, RSG \citep{robotskillgraph}, and \frame) manage to incorporate \expsim{} also during the learning (i.e., lifelong learning of \expsim). HlifeRL \citep{HlifeRL_optionbased} learns to combine different skill options using simulation and task context provided; consequently, it cannot operate autonomously without human intervention. Similarly, RSG \citep{robotskillgraph} constructs an initial skill graph using large and diverse fundamental skills of up to 320 skills, which are pretrained in simulation. 

In contrast to those previous works, \frame{} demonstrates the utilization of interpretable neural control structure in distributing the functions to different components, enabling the sharing of learned skills during both \init{} and learning while preventing the robot-skill encoding parameters from catastrophic forgetting under autonomous lifelong learning directly in the real world. Therefore, to the best of our knowledge, \frame{} is the only locomotion learning framework that realizes online continual locomotion learning in the real world without catastrophic forgetting while exploiting task similarity during both the \init{} and learning phases without task context or human intervention (under unlimited space).

\subsubsection{Interpretability Challenge} 

The last challenge is the inability to understand and verify the neural controllers owing to their black-box nature. To deal with this, \frame{} is designed to include three characteristics: decomposability, transparency, and simulability \citep{review_mythos,taxonomiesXAI,reviewInterpretableRL}. \frame{} uses interpretable modular layers and column-wise subnetworks to enhance its decomposability, assign distinct functions to individual parameters, provide a transparent learning process, and enable simulability by allowing the conversion between neural control networks and \behav s, which present complex behavior organization and transitions. These characteristics are supported by quantitative assessments using compactness and completeness \citep{evaluationXAI2}, which reveal that \frame{} is simpler to interpret (due to its compactness) without significantly compromising fidelity (due to its completeness).

\subsection{Engineering Aspect}

Concerning the engineering aspect, \frame{} can be applied for developing robot locomotion in different unseen condition, including energy efficient gaits on different terrains \citep{hexapod_optimize_cot}, changing slope \citep{zumogecko}, and abnormal conditions \citep{hexapod_legamp} or be employed for developing simple unsupervised decision making, e.g., to classify terrains \citep{hexapod_blindlearning,hexapod_terrainclassification2} unsupervisedly without true targets. Additionally, \frame{} provide an option for developing various behaviors which are self-organized in an interpretable hierarchical structure, resembling a state machine, a behavior tree, or a motion primitive graph \citep{behaviortree_statemachine,demonstration_motionprimitivegraph}, which human can understand and modify. This developed behavior hierarchy can also be interpreted as a map, inferring the robot path and the structure of the learning environment. Lastly, all these are not limited to solely to hexapod robot locomotion as it can be applied to other types of robot, e.g., a hexapod robot with amputated middle legs (in the \sup{} and \videoquad) and a quadruped robot (in the \sup{} and \videobone) by only changing the output dimension from 18 joints to 12 joints. Moreover, \frame{} can be extended to other domains, such as, programming by demonstration. This can be achieve by only replacing the locomotion reward (locomotion pattern mapping) and value prediction (neurogenesis) with a fitting error function (supervised learning, see \sup). The core neural control structure (Figure~\ref{fig:method}), however, remains unchanged. This \frame-based programming-by-demonstration method has been applied to robot arm manipulation involving action sequences (see \videodemo). Additionally, it can be also used to program hexapod leg motion through kinesthetic demonstration (see  \videodemomorf).

\subsection{Bio-inspired Robotics and Robotics-inspired Biology Aspects}

Regarding the bio-inspired robotics and robotics-inspired biology aspects, \frame{} exhibits several bio-inspired life long learning features \citep{bio_liftlonglearning}, as summarized in the \sup. In addition, \frame{} constitutes another possible supporting model/hypothesis for future biology research and robotics-inspired biology \citep{robotinspiredbio}. Apart from providing a supporting mechanism for life long learning \citep{bio_liftlonglearning,lifelong_hamburg}, neural control exhibits the combination of feedback independent activity propagation for rhythmic generation, such as central pattern generators \citep{maincpg}, and feedback dependent activity propagation for conditional process, such as decision making in humans \citep{bio_reaching_actionbased_model1,bio_reaching_actionbased_model2} and animals \citep{nematodedecision} as well as perception-like orientation estimation in \textit{Drosophila melanogaster} \citep{bio_ringattractor}. Additionally, the adaptation of the exploration rate based on reward/penalty as well as learning prediction uncertainty boundaries with a slow learning rate in \frame{} could be analogies and possible mathematical models for studying biological adaptation signals, such as neuromodulation \citep{acl1,learninghormone}.

Ultimately, \frame{} could pave the way for fully autonomous lifelong learning machines with motion intelligence that can adapt and learn in diverse conditions without human intervention. This could introduce a new form of collaboration between humans and robots where individuals can actively cooperate with these lifelong learning machines by interpreting their learning results and introducing them to new environments for further learning. Moreover, it could redefine our role, shifting from extensive robot programming to active robot supervision, empowering individuals to train and engage with their own robots regardless of their robotics knowledge and further contributing to the rapid development of robotics technologies.

\section{Acknowledgments} 
\label{sec:ack}
This work was supported by the fellowship from Vidyasirimedhi Institute of Science and Technology and by the Marie Sk\l{}odowska-Curie Actions-Doctoral Networks grant agreement No 101119614 (MAESTRI) (PM, WP6-PI). In addition, we thank Kasper St\o y and Chaicharn Akkawutvanich for discussions and suggestions, Cao Dah Do and Atthanat Harnkhamen for the hardware and experimental support, \reviewfinal{and Senior Editor Prof. Huaping Liu and Associate Editor Dr. Danfei Xu for their supervision of the review process.}

\section{Supplementary Materials}
\begin{itemize}
	\itemindent=-10pt
	\item Program: \gitlink
	\item Interpretable Neural Locomotion Control: \videodecompose{} and \videobehavmodel
	\item Locomotion learning on flat terrain: \videonormal
	\item Continual locomotion learning on different slopes: \videoslope
	\item Continual locomotion learning on different slopes with possible motor dysfunction: \videobroken
	\item Continual locomotion learning on different terrains: \videoterrain
	\item Quadruped locomotion learning: \videoquad{} (amputated MORF) and \videobone{} (quadruped robot)
	\item Programming by demonstration: \videodemo{} and \videodemomorf
	
\end{itemize}

\bibliographystyle{SageH}
\bibliography{ref.bib}

\if0
\pagebreak
\clearpage
\onecolumn
\begin{center}
	\textbf{\large Supplementary Material}
\end{center}
\setcounter{equation}{0}
\setcounter{figure}{0}
\setcounter{table}{0}
\setcounter{page}{1}
\setcounter{section}{0}
\makeatletter
\renewcommand{\theequation}{S\arabic{equation}}
\renewcommand{\thefigure}{S\arabic{figure}}
\renewcommand{\thesection}{S\arabic{section}}
\renewcommand{\thetable}{S\arabic{table}}
\input{section/supplementary}
\fi

\end{document}